%% file: main.tex
\documentclass[10pt,twocolumn,letterpaper]{article}

\usepackage[pagenumbers]{wacv} 

\input{preamble}

\usepackage[pagebackref,breaklinks,colorlinks]{hyperref}

\usepackage[capitalize]{cleveref}
\crefname{section}{Sec.}{Secs.}
\Crefname{section}{Section}{Sections}
\Crefname{table}{Table}{Tables}
\crefname{table}{Tab.}{Tabs.}

\def\wacvPaperID{0000} \def\confName{WACV}
\def\confYear{2025}

\makeatletter
\renewcommand{\paragraph}{\@startsection{paragraph}{4}{\z@}{2.0ex \@plus 1ex \@minus .2ex}{-1em}{\normalfont\normalsize\bfseries}}
\makeatother

\newcommand{\PAR}[1]{\vskip3pt \noindent {\bf #1~}}

\begin{document}

\title{Fine-Tuning Image-Conditional Diffusion Models is Easier than You Think}

\author{Gonzalo Martin Garcia$^1$\qquad 
Karim Abou Zeid$^1$\qquad
Christian Schmidt$^1$\\
Daan de Geus$^{1,2}$\qquad
Alexander Hermans$^1$\qquad
Bastian Leibe$^1$\\[0.6em]
$^1$RWTH Aachen University\qquad $^2$Eindhoven University of Technology\\[0.6em]
\httpsurl{vision.rwth-aachen.de/diffusion-e2e-ft}
}

\twocolumn[{\renewcommand\twocolumn[1][]{#1}\maketitle
\newcommand{\rowheight}[0]{2.6cm}
\newcommand{\figa}[0]{lego2}
\newcommand{\figb}[0]{statue2}
\newcommand{\figc}[0]{chess}
\newcommand{\figd}[0]{museum}
\setlength{\tabcolsep}{1pt}
\begin{tabular}{cccc}
 \includegraphics[height=\rowheight]{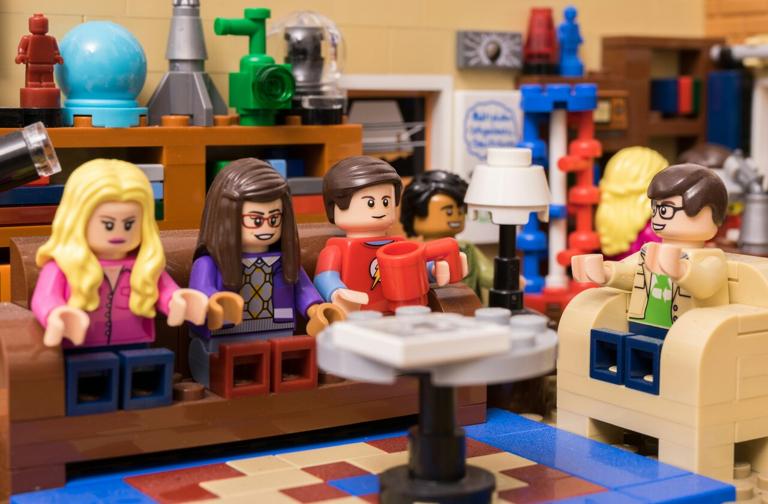}&\includegraphics[height=\rowheight]{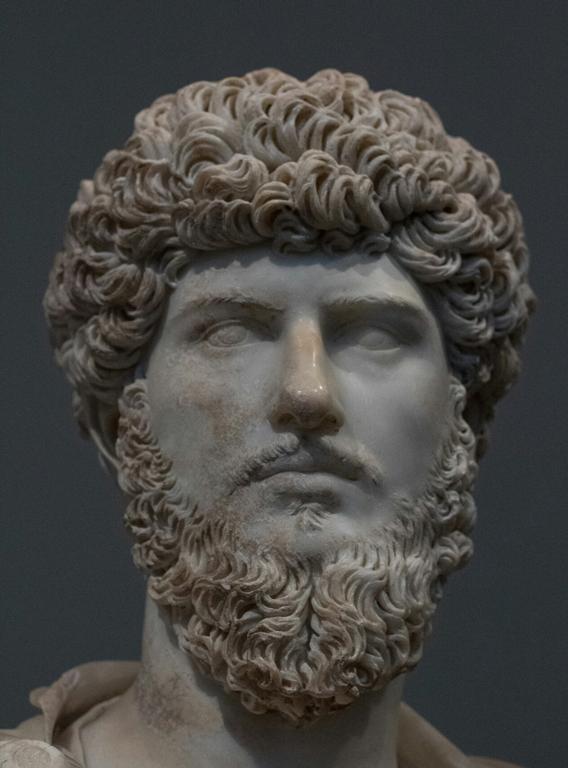}&\includegraphics[height=\rowheight]{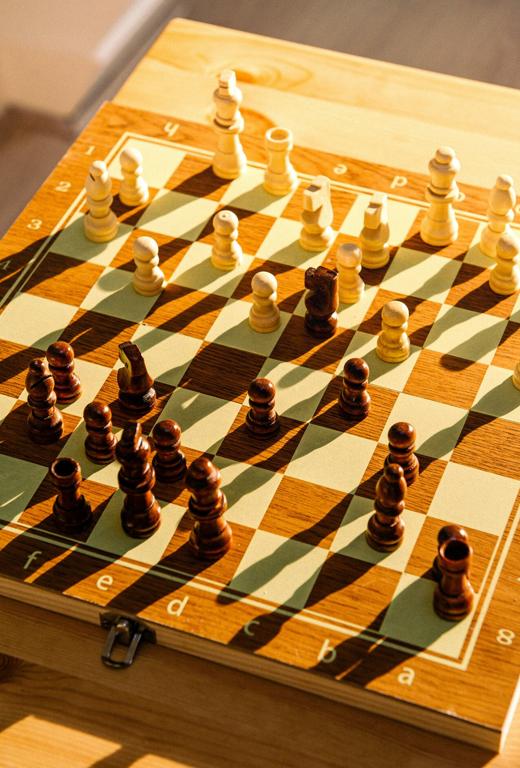}&\includegraphics[height=\rowheight]{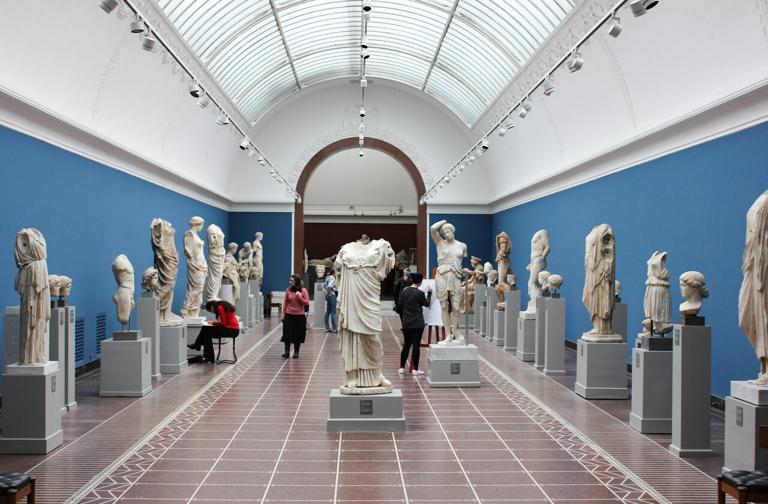}\\[-1pt]\includegraphics[height=\rowheight]{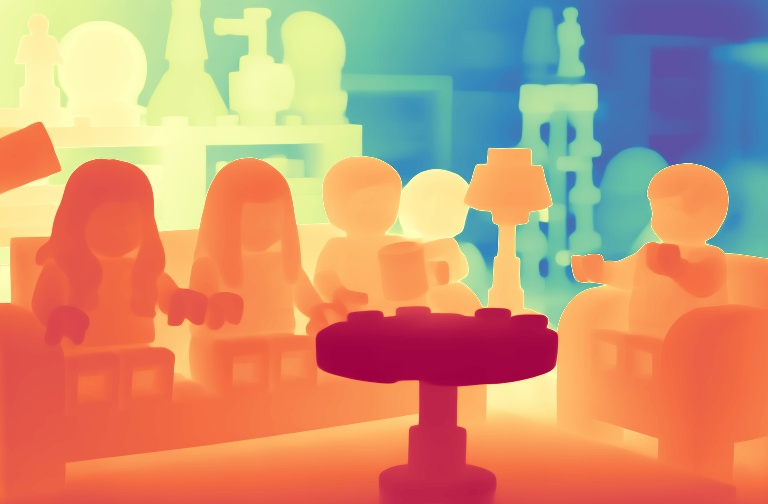}&\includegraphics[height=\rowheight]{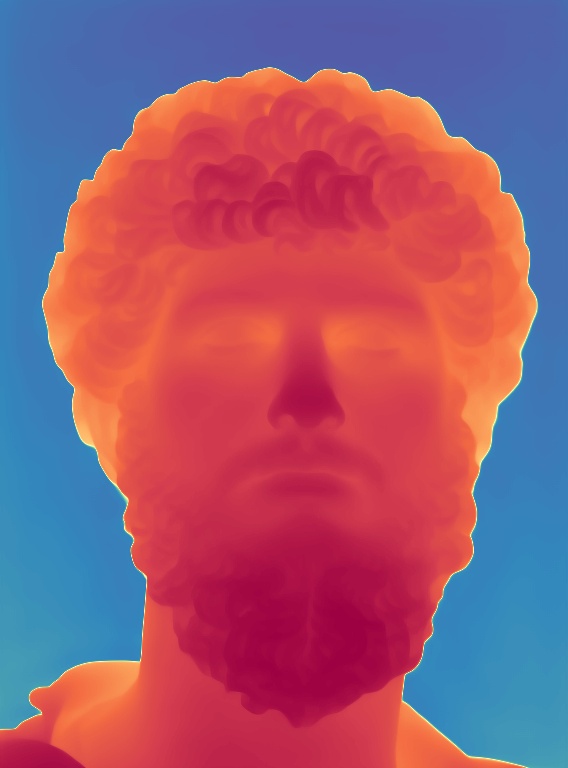}&\includegraphics[height=\rowheight]{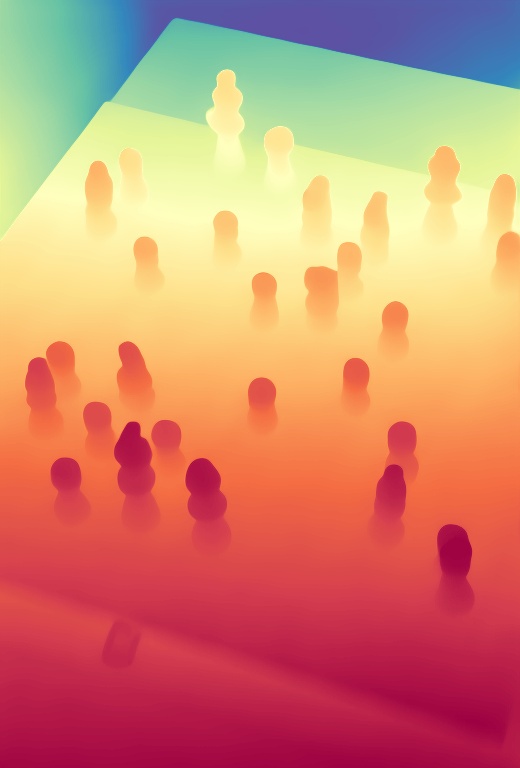}&\includegraphics[height=\rowheight]{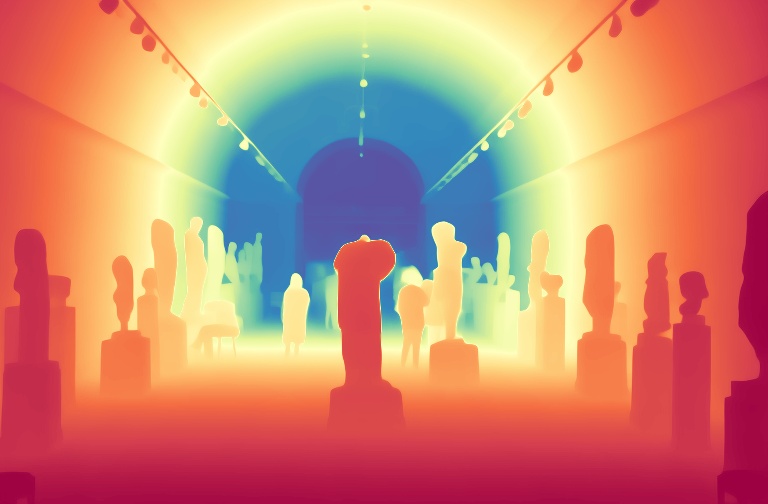}\\[-1pt]\includegraphics[height=\rowheight]{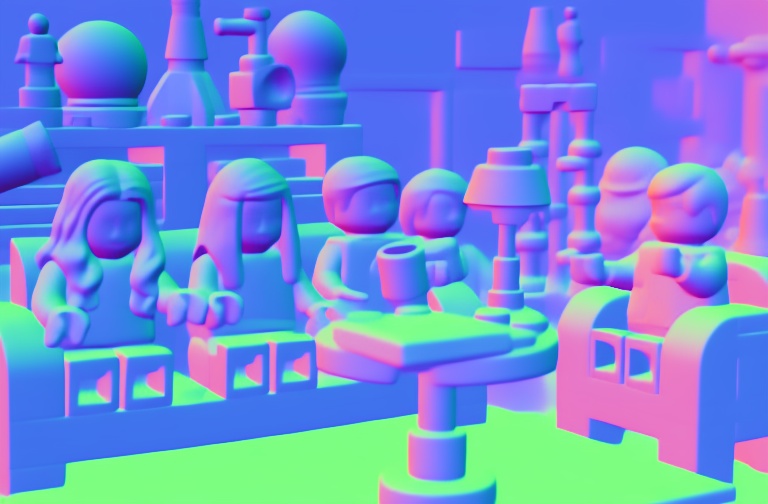}&\includegraphics[height=\rowheight]{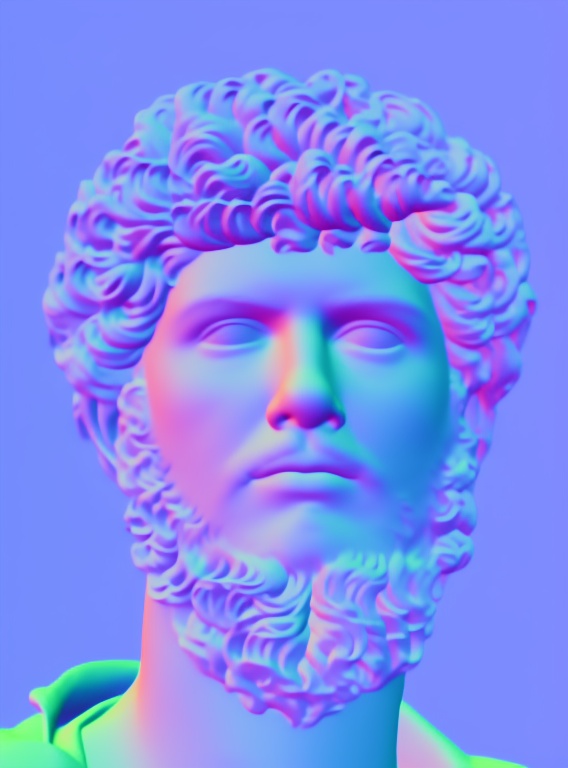}&\includegraphics[height=\rowheight]{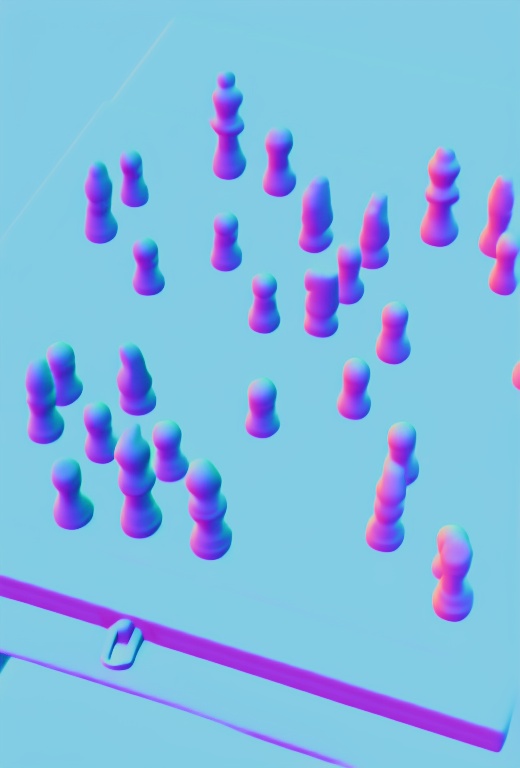}&\includegraphics[height=\rowheight]{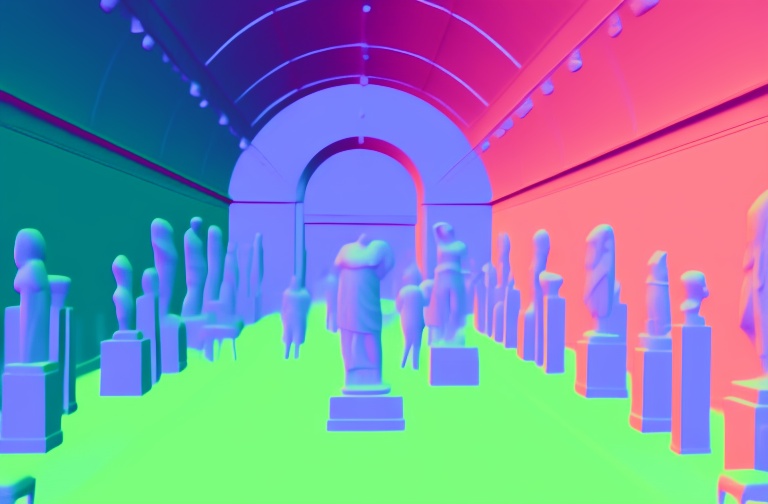}\end{tabular}~~~\raisebox{-3.95cm}{\input{figures/teaser_plot}}
\vspace{-6pt}
\captionof{figure}{
    \textbf{Repurposing diffusion models for geometry estimation is as simple as end-to-end fine-tuning.}
    Left: Depth and normal predictions of our method on in-the-wild images.
    Right: A simple fix for the DDIM scheduler enables single-step inference for recent diffusion-based depth estimators; and simple end-to-end fine-tuning outperforms more complex diffusion baselines in speed and accuracy.
\vspace{0.5em}
}
\label{fig:teaser}
}]

\setlength{\abovecaptionskip}{3pt plus 4pt minus 2pt}
\setlength{\abovedisplayskip}{6pt}
\setlength{\belowdisplayskip}{6pt}
\setlength\abovedisplayshortskip{6pt}
\setlength\belowdisplayshortskip{6pt}

\aboverulesep = 0.4mm
\belowrulesep = 0.5mm

\setlength{\textfloatsep}{7pt plus3pt minus3pt}
\setlength{\dbltextfloatsep}{7pt plus3pt minus3pt}

\input{sec/0_abstract}    
\input{sec/1_intro}
\input{sec/2_related_work}

\input{sec/3_background}
\input{sec/4_method}
\input{sec/5_experiments}

\input{sec/6_discussion}

{\small
\bibliographystyle{ieee_fullname}
\bibliography{egbib}
}

\clearpage
\input{sec/X_suppl}

\end{document}

%% file: preamble.tex
\usepackage{graphicx}
\usepackage{amsmath}
\usepackage{amssymb}
\usepackage{booktabs}
\usepackage{siunitx}
\usepackage{bm}
\usepackage{adjustbox}
\usepackage{csquotes}
\usepackage[normalem]{ulem}

\usepackage[svgnames, table]{xcolor}

\definecolor{noreproduce}{gray}{0.6}

\usepackage{multirow}

\usepackage{makecell}
\usepackage{tabularx}
\newcolumntype{Y}{>{\centering\arraybackslash}X}

\newcommand{\I}{\mathbf{I}}

\newcommand{\vect}[1]{\boldsymbol{\mathbf{#1}}}

\usepackage{tikz}
\usetikzlibrary{calc}
\usetikzlibrary{backgrounds}

\newcommand{\downrightarrow}{\hspace{3pt}\raisebox{1.5pt}{\begin{tikzpicture}[scale=0.4, baseline=(current bounding box.south)]
    \draw[-latex] (0.0, 0.5) -- (0.0, 0.25) -- (0.7, 0.25);
  \end{tikzpicture}} }

\newcommand\ExtraSep
{\dimexpr\cmidrulewidth/2+\aboverulesep/2+\belowrulesep/2\relax}

\newcommand{\mr}[2][c]{\multirowcell{2}[-\ExtraSep][#1]{#2}}

\newcommand{\httpsurl}[1]{\href{https://#1}{\texttt{#1}}}

\makeatletter
\newcommand\notsotiny{\@setfontsize\notsotiny{5.5}{6.5}}
\makeatother

\newcommand{\ven}[1]{\tiny\color{gray}(#1)}

\usepackage{subcaption}

\usepackage{pgfplots}
\usepackage{pgfplotstable}
\pgfplotsset{compat=1.18}
\usepgfplotslibrary{groupplots} 

\definecolor{marigold_col}{RGB}{219, 34, 34}
\definecolor{marigold10_col}{RGB}{248, 145, 144}
\definecolor{marigold_fixed_col}{RGB}{34, 108, 167}
\definecolor{marigold10_fixed_col}{RGB}{158, 199, 221}
\definecolor{ours_col}{RGB}{65,200,35}

\usetikzlibrary{shapes.geometric}
\pgfdeclareplotmark{mystar}{
    \node[star, star points=5, star point ratio=0.5, rotate=180, draw=ours_col!90!black, solid, fill=ours_col, minimum width=3pt, inner sep=0pt, outer sep=0pt, anchor=center] {};
}

\usepackage{float}

%% file: figures/teaser_plot.tex
\begin{tikzpicture}[tight background, font=\footnotesize]
    \begin{axis}[
        xmode=log,
        xmin=0.8,
        xmax=600,
        xtick={1,2,4,10,25,50,150,500},
        xticklabels={1,2,4,10,25,50,150,500},
        ymin=4.8,
        ymax=36,
        ymode=log,
        ytick={5,6,7,10,16,30},
        yticklabels={5.0,6.0,7.0, 10.0, 16.0, 30.0},
        xlabel={Neural Function Evaluations ↓},
        ylabel={AbsRel ↓~~~~~},
        legend style={
            at={(0.95,0.9)},
            anchor=north east,
            legend columns=1,
            legend cell align={left},
            draw=gray!30, line width=0.8pt,
            inner ysep=2pt,
            inner xsep=3pt,
        },
        grid=both,
        tick style={draw=none}, major grid style={line width=0.8pt,draw=gray!30},
        axis line style={draw=gray!30, line width=0.8pt},
        width=5.8cm,
        height=8.8cm,
        xlabel shift={-2pt},
        ylabel shift={-9pt},
    ]
    \addplot[color=marigold_col, dotted, thick, mark=*, mark options={fill=marigold_col, solid}] table[x=steps, y= mari_absrel, col sep=comma] {data/figure_data.csv};

    \addplot[color=marigold_col, dotted, thick, mark=*, mark options={fill=marigold_col, solid}] table[x expr=\thisrow{steps}*10, y=mari10_absrel, col sep=comma, forget plot] {data/figure_data.csv};
    \addlegendentry{Marigold};

    \addplot[color=marigold_fixed_col, dotted, thick, mark=*, mark options={fill=marigold_fixed_col, solid}] table[x=steps, y= marifix_absrel, col sep=comma, forget plot] {data/figure_data.csv};

    \addplot[color=marigold_fixed_col, dotted, thick, mark=*, mark options={fill=marigold_fixed_col, solid}] table[x expr=\thisrow{steps}*10, y=mari10fix_absrel, col sep=comma] {data/figure_data.csv};
    \addlegendentry{Marigold (fixed)};

\pgfplotstableread[col sep=comma]{data/figure_data.csv}\loadedtable
    \pgfplotstablegetelem{0}{ours_absrel}\of\loadedtable
    \let\yvalue\pgfplotsretval
\addplot [mark=mystar, only marks] coordinates {(1, \yvalue)};
    \addlegendentry{Marigold + E2E FT};
    \addplot [ours_col, dashed, thick, domain=0.1:600] {\yvalue};
    
    \end{axis}
    \end{tikzpicture}

%% file: sec/0_abstract.tex
\begin{abstract}
Recent work showed that large diffusion models can be reused as highly precise monocular depth estimators by casting depth estimation as an image-conditional image generation task.
While the proposed model achieved state-of-the-art results, high computational demands due to multi-step inference limited its use in many scenarios.
In this paper, we show that the perceived inefficiency was caused by a flaw in the inference pipeline that has so far gone unnoticed.
The fixed model performs comparably to the best previously reported configuration while being more than 200$\times$ faster.
To optimize for downstream task performance, we perform end-to-end fine-tuning on top of the single-step model with task-specific losses and get a deterministic model that outperforms all other diffusion-based depth and normal estimation models on common zero-shot benchmarks.
We surprisingly find that this fine-tuning protocol also works directly on Stable Diffusion and achieves comparable performance to current state-of-the-art diffusion-based depth and normal estimation models, calling into question some of the conclusions drawn from prior works.
\end{abstract}

%% file: sec/1_intro.tex
\section{Introduction}
\label{sec:intro}

Monocular depth estimation has long been used in many downstream tasks, such as image and video editing, scene reconstruction, novel view synthesis, and robotic navigation. 
Since the task is inherently ill-posed due to the scale-distance ambiguity, learning-based methods need to incorporate strong semantic priors in order to perform well.
For this reason, recent work has proposed to adapt large diffusion models~\cite{rombach2021stablediffusion2} for monocular depth estimation by casting depth prediction as a conditional image generation task~\cite{ke2023marigold}.
The resulting models show good task performance and exhibit remarkably high levels of details.
However, the consensus in the community is that they tend to be slow~\cite{ke2023marigold,fu2024geowizard,gui2024depthfm}, since they need to perform many evaluations of a large neural network during inference.

In this paper, we argue that, contrary to common belief, inference of conditional latent diffusion models such as Marigold~\cite{ke2023marigold} and follow-up work~\cite{fu2024geowizard} should be able to yield reasonable predictions with a single inference step.
We investigate the behavior of Marigold and find that its dismal performance in the few-step regime is due to a critical flaw in the inference pipeline.
While this bug has already been reported in the general diffusion model literature~\cite{Lin2023CommonDiffusionFlawed}, we demonstrate that it is particularly critical in the scope of image-conditional methods such as Marigold.
In particular, our results indicate that \emph{existing works have probably drawn wrong conclusions due to flawed inference results}.

With a small correction to the inference pipeline, Marigold-like models~\cite{ke2023marigold,fu2024geowizard} obtain single-step performance that is comparable to multi-step, ensembled inference, while \textbf{being more than 200$\times$ faster}.
In fact, this bug-fix makes diffusion-based depth estimators speed-wise comparable to state-of-the-art discriminative depth estimation models, opening up exciting avenues for further improvements.
First, a single-step model allows efficient task-specific end-to-end fine-tuning since there is no need to backpropagate through multiple network invocations.
Second, advanced techniques such as self-training with pseudo-labels~\cite{yang2024depthanything,yang2024depthanythingv2}, which have been proven to be effective for discriminative models, can now be efficiently applied to diffusion-pretrained models as well.

We fine-tune Marigold end-to-end into a deterministic affine-invariant depth estimator for monocular images using a scale and shift invariant loss function~\cite{Ranftl2022midas}.
To our surprise, this model outperforms the best
configurations of Marigold.
We repeat this experiment with the task of surface normal estimation and find similar results: end-to-end fine-tuning with a task-specific loss outperforms more complicated architectures which were trained on more data.

Following Occam's Razor, we find that even the simplest baseline, direct fine-tuning of Stable Diffusion (SD)~\cite{rombach2021stablediffusion2} into a deterministic feed-forward model, outperforms Marigold and other diffusion-based depth- and normal estimation methods.
These findings contradict some conclusions that have been drawn in earlier works.
First, diffusion-based depth and normal estimation methods do not need to be slow.
Second, casting depth estimation as conditional image generation is not more effective than simple end-to-end fine-tuning.
But in line with existing intuitions, we find a small dataset of high-quality (synthetic) labeled data sufficient for good performance.

In summary, our contributions are:
(1) we analyze the behavior of Marigold and similar diffusion-based geometry estimation models and find a critical flaw in their inference pipeline, (2) we fix this critical flaw, enabling high-precision single-step inference and boosting efficiency of these models by more than 200$\times$, and
(3) we show that simple task-specific fine-tuning of diffusion models is sufficient for good performance in depth and normal estimation.

We demonstrate the effectiveness of our approach on common zero-shot benchmarks.
Our deterministic one-step model outperforms other diffusion-based depth- and normal estimation methods, and achieves results comparable to state-of-the-art methods for affine-invariant depth prediction and surface normal estimation.

%% file: sec/2_related_work.tex
\section{Related Work}
\label{sec:related_work}
\PAR{Monocular Depth Estimation.}
Monocular depth estimation models predict a pixel-wise depth map of a scene from a single image.
The most comprehensive representation is \emph{metric depth}, which requires modeling the focal length to account for different cameras, introducing additional uncertainty~\cite{bhat2023zoedepth,piccinelli2024unidepth,yin2023metric3d,hu2024metric3dv2}.

An alternative to metric depth is \emph{affine-invariant depth}, which is equivalent to metric depth up to an unknown global scale and shift of the scene.
This is the representation of choice for a wide range of monocular depth estimation methods~\cite{zhang2022hdn,Ranftl2022midas,ranftl2021dpt,eftekhar2021omnidata,kar2022omnidatav2,yang2024depthanything,yang2024depthanythingv2}.
Unlike metric depth, affine-invariant depth is independent of the focal length and is easier to regress in unfamiliar scenes, where no object can serve as a metric reference.
However, it still preserves the distance ratios between objects.
Metric depth can be recovered from affine-invariant depth by anchoring it with sparse, known depth values or by explicitly estimating the missing scale and shift~\cite{yin2021leres}.

To generalize to \enquote{in-the-wild} unseen scenes, depth estimation methods must handle a wide variety of environments.
Methods designed for such applications are typically evaluated in a zero-shot setting, where the model is tested on unseen datasets without fine-tuning.
Early zero-shot depth estimation methods already focused on generalization primarily through (at the time) large training datasets~\cite{li8megadepth,yin2021diversedepth}.
MiDaS~\cite{Ranftl2022midas} achieved significant improvements by leveraging a combination of multiple datasets and a high-capacity backbone.

The transition from CNNs to ViTs~\cite{dosovitskiy2020vit} in DPT~\cite{ranftl2021dpt} and Omnidata~\cite{eftekhar2021omnidata} marked another key advancement in the field.
Recent methods such as Depth Anything~\cite{yang2024depthanything,yang2024depthanythingv2} and Metric3D~\cite{yin2023metric3d,hu2024metric3dv2} have followed the success of MiDaS by utilizing the high-capacity ViT-g DINOv2~\cite{oquab2023dinov2} backbone and training on vast datasets---62M and 16M samples, respectively---one to two orders of magnitude larger than previous work.
Notably, Depth Anything retains a simple DPT architecture, but combining a DINOv2 initialization and a large training dataset enables it to generalize to in-the-wild scenarios.
Metric3D additionally utilizes the focal length to boost performance, which however limits its training data to scenes with known focal length.

\PAR{Monocular Normal Estimation.}
Surface normal estimation involves predicting the orientation of surfaces in a scene from an image, resulting in a 3D vector representing the surface's orientation for each pixel.
Hoiem \etal~\cite{Hoiem2005AutomaticPP,Hoiem2007RecoveringSL} were among the pioneers to introduce learning-based approaches for surface normal estimation. Since then, deep learning methods have become dominant, with many notable contributions~\cite{Wang2014DesigningDN,eigen2015predictingdepth,Bansal2016MarrR2,Bae2021eesnu,eftekhar2021omnidata,kar2022omnidatav2,bae2024dsine}.
Among these, Omnidata~\cite{eftekhar2021omnidata} stands out for training a UNet-based model on a large-scale dataset of 12M images captured in diverse environments. Its successor, Omnidata v2~\cite{kar2022omnidatav2}, advanced the field by transitioning to a ViT-based architecture with a DPT head, incorporating sophisticated 3D-aware data augmentations to enhance generalization. In contrast, DSINE~\cite{bae2024dsine} adopts a more data-efficient approach that focuses on introducing new inductive biases to enhance performance. Lastly, Metric3D v2~\cite{hu2024metric3dv2}, mentioned earlier, is also capable of predicting surface normals in addition to depth.

\PAR{Diffusion Models for Geometry Estimation.}
Several recent generative text-to-3D methods~\cite{long2024wonder3d,qiu2024richdreamer} explicitly produce multi-view depth and normal maps.
However, these methods focus on isolated single-object scenarios, making them unsuitable for complex, in-the-wild environments.

Other approaches have utilized diffusion models for scene-level depth estimation~\cite{saxena2023depthgen,saxena2023ddvm,duan2023diffusiondepth,zhao2023VPD}.
Among these, VPD~\cite{zhao2023VPD} leverages Stable Diffusion~\cite{rombach2021stablediffusion2} as both an image and a text feature extractor, incorporating a depth regression head that utilizes multi-scale image feature maps alongside text-to-image cross-attention maps.
However, these models have not demonstrated robust generalization.

More recently, Marigold~\cite{ke2023marigold} fine-tuned SD to transition from generating realistic images conditioned on text to producing detailed and precise depth maps conditioned on images.
The core idea behind this approach is that SD's ability to model realistic images also provides strong geometric and semantic priors, essential for accurate depth estimation.
Key to Marigold's success is its exclusive training on high-quality synthetic datasets with perfect ground truth and a smooth transition from text-conditioned images to image-conditioned depth in the latent space, preserving the model's generalization capability.

Marigold has inspired several follow-up works.
DiffCalib~\cite{he2024diffcalib}, for example, extends Marigold by jointly predicting depth and camera intrinsics through the addition of an incident map~\cite{zhu2023tamewildcamera}, which is denoised alongside the depth map.
GeoWizard~\cite{fu2024geowizard} jointly predicts both depth and surface normals through two parallel UNet evaluations, incorporating cross-attention between the two branches.
However, a common drawback of these models is the high computational cost during inference, driven by the requirement for an iterative denoising process.

Addressing this limitation, DepthFM~\cite{gui2024depthfm} combines Marigold's core ideas with Flow Matching~\cite{lipman2022flowmatching} to reduce the number of denoising steps while maintaining high output quality.
Additionally, the authors of Marigold now provide an LCM-distilled~\cite{luo2023LCM} version that allows for single-step evaluation, albeit at a reduced quality.

In our work, we observe that Marigold and follow-up methods, aside from DepthFM and Marigold LCM, which are designed for few-step prediction, suffer from a flawed implementation~\cite{Lin2023CommonDiffusionFlawed} of the DDIM~\cite{song2020DDIM} inference pipeline that prevents them from functioning effectively in the few-step regime.
Furthermore, although the denoising diffusion fine-tuning objective used by Marigold and follow-up works for depth and normals estimation has shown effectiveness, we find it to be neither a key factor for good results nor clearly superior to task-specific end-to-end fine-tuning.

%% file: sec/3_background.tex
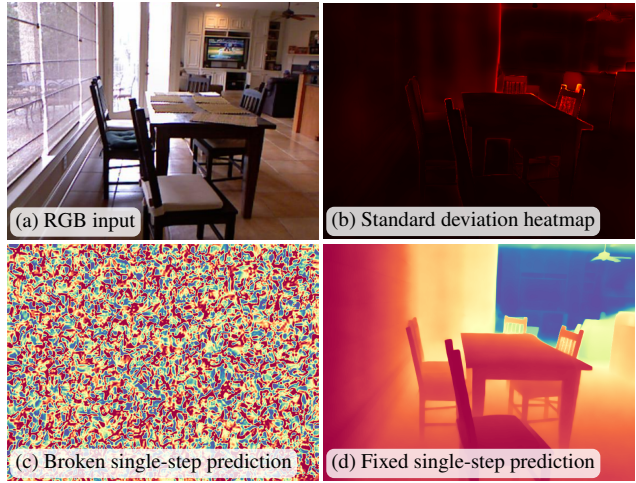
\begin{figure}
    \centering
    \begin{adjustbox}{max width=\linewidth}
    \begin{tikzpicture}[
        image/.style={
            inner sep=0pt,
            outer sep=1pt,
        },
        label/.style={
            anchor=south west, 
            outer sep=2pt,
            inner sep=2pt,
            color=black, 
            fill=white,
            opacity=0.8,
            text opacity=1,
            rounded corners=3pt,            
        }
    ]
    \node[image] (rgb) at (0,0) {\includegraphics[width=0.5\linewidth]{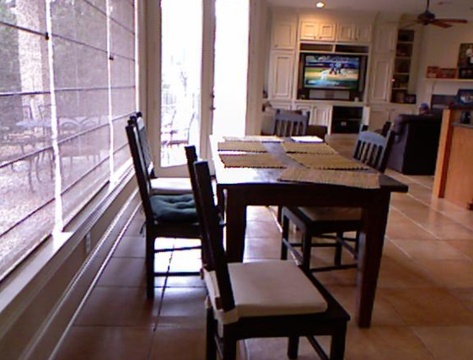}};
    \node[label] (rgb_text) at (rgb.south west) {
        \footnotesize (a) RGB input\phantomsubcaption\label{fig:comparison_heatmap_rgb}
    };
    \node[image, anchor=north west] (heatmap) at (rgb.north east) {\includegraphics[width=0.5\linewidth]{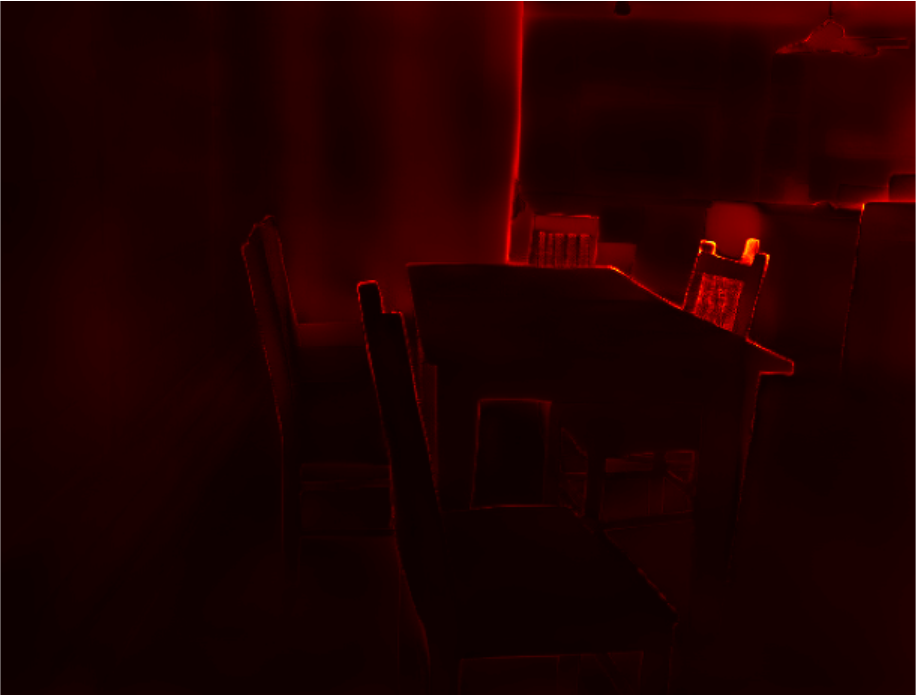}};
    \node[label] (heatmap_text) at (heatmap.south west) {
        \footnotesize (b) Standard deviation heatmap\phantomsubcaption\label{fig:comparison_heatmap_std}
    };
    \node[image, anchor = north west] (maribroke) at (rgb.south west) {\includegraphics[width=0.5\linewidth]{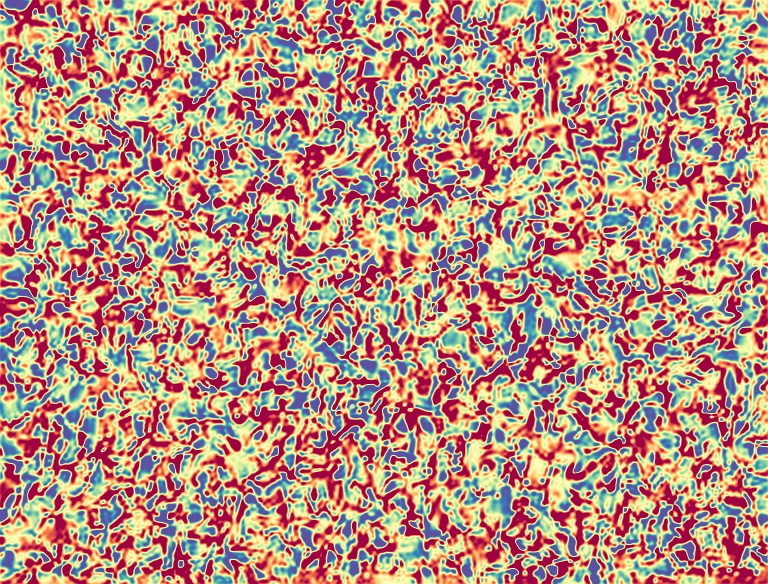}};
    \node[label] (maribroke_text) at (maribroke.south west) {
        \footnotesize (c) Broken single-step prediction\phantomsubcaption\label{fig:comparison_heatmap_maribroke}
    };
    \node[image, anchor = north west] (marifixed) at (rgb.south east){\includegraphics[width=0.5\linewidth]{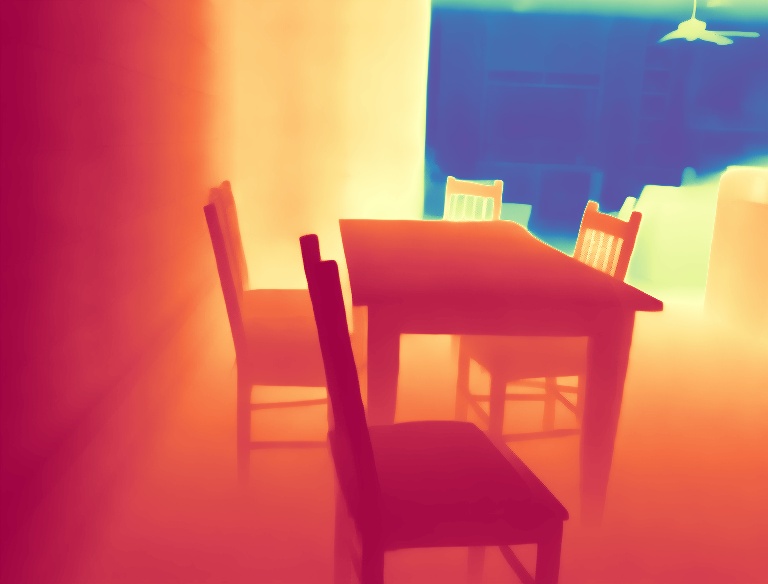}};
    \node[label] (marifixed_text) at (marifixed.south west) {
        \footnotesize (d) Fixed single-step prediction\phantomsubcaption\label{fig:comparison_heatmap_marifixed}
    };
    \end{tikzpicture}
    \end{adjustbox}
    \caption{
        \textbf{Marigold output visualizations.} (a) The RGB input; (b) the pixel-wise standard deviation of Marigold's depth map output during 50-step DDIM inference; and Marigold's depth map prediction (c) before and (d) after the fixing the inference pipeline.
    }
    \label{fig:comparison_heatmap}
\end{figure}

\section{Image-Conditional Latent Diffusion Models}
\label{sec:background}

In this section, we review conditional latent diffusion models and how Marigold~\cite{ke2023marigold} leverages them for depth estimation.
We also argue why single-step inference should produce sensible predictions for depth estimation, and explain why this has not been the case in practice so far.

\subsection{Latent Diffusion Models}
\label{ssec:background_ldm}
\emph{Denoising Diffusion Probabilistic Models (DDPM)} learn a mapping from some simple, known noise distribution $p_T$ to the data distribution $p_0$ by reversing a stochastic forward process $p_t,\ t=1,\ldots,T$, which repeatedly adds a small amount of Gaussian noise~\cite{ho2020ddpm}.
The variance $\beta_t$ of the noise added in each step is chosen to be small, so that the reverse process can be approximated as Gaussians.
Additionally, the number of steps $T$ is set sufficiently large such that the terminal distribution $p_T$ can be approximated as a Gaussian as well.
Using $\alpha_t = 1 - \beta_t$ and $\bar\alpha_t = \prod_{\tau=1}^t \alpha_\tau$, the forward process can be written as $\vect{x}_t = \sqrt{\bar\alpha_t}\vect{x}_0 + \sqrt{1 - \bar\alpha_t} \vect{\epsilon}$ given a data sample $\vect{x}_0$ and $\vect{\epsilon} \sim \mathcal{N}(\textbf{0}, \I)$.
For the reverse process, a neural network is trained to gradually remove noise from its inputs to predict $\vect{x}_{t-1}$ given $\vect{x}_t$~\cite{ho2020ddpm}.

Inference starts with noise $\vect{x}_T$, which is repeatedly denoised by passing it through the diffusion model and thus following the reverse diffusion process.
A popular alternative inference scheme are \emph{Denoising Diffusion Implicit Models (DDIM)}~\cite{song2020DDIM}, which formulate a non-Markovian diffusion process that leads to the same training objective as DDPMs, but allows for inference in just a few steps.

\emph{Latent Diffusion Models (LDMs)}\cite{rombach2021stablediffusion2} operate in the latent space of another model, \eg, a \emph{Variational Autoencoder (VAE)}~\cite{kingma2013vae}.
The VAE consists of an encoder $\mathcal{E}$ and a decoder $\mathcal{D}$ and is trained independently of the LDM.
The goal of the VAE encoder is to compress inputs into lower-dimensional latent codes, and for the decoder to faithfully reconstruct the input: $\mathcal{D}(\mathcal{E}(\vect{x}))\approx\vect{x}$.
LDMs tend to be easier to train due to the reduced dimensionality and improved smoothness of the VAE's latent space.

In conditional diffusion models, both forward and reverse processes are conditioned on some additional input $\vect{c}$, such as a text description or an additional image input.
The network is then trained to denoise the input given $\vect{c}$.

It has been shown that the optimal prediction at the final timestep $T$ is the mean of the data distribution~\cite{karras2022elucidating}.
For unconditional and text-conditional image generation, the data distribution has multiple modes, and the mean therefore corresponds to a blurry image dominated by the average color and average scene composition.
For a text-conditional model, the mean of the image distribution conditioned on the input text will be a blurry resemblance of the average image fitting the description.
But for image-conditional generation, in particular depth and normal estimation, we expect the conditional distribution to be approximately unimodal, since an image usually corresponds to a single depth map.
The optimal single-step prediction at $T$ should therefore be close to the ground-truth depth/normal map.

\subsection{Marigold}
\label{ssec:background_marigold}

\begin{figure}
    \centering
    \includegraphics[width=\linewidth]{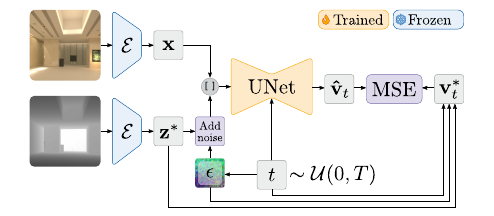}
    \caption{
        \textbf{Marigold~\cite{ke2023marigold} diffusion training for conditional depth map generation.}
        Marigold starts with a pretrained Stable Diffusion v2 model~\cite{rombach2021stablediffusion2}, which is fine-tuned for image-conditional generation of depth maps or surface normal maps.
    }
    \label{fig:overview_training}
\end{figure}

Marigold casts depth estimation as a conditional latent diffusion process, allowing it to build upon large pretrained diffusion models such as Stable Diffusion~\cite{ke2023marigold}.
Marigold is conditioned on images, so following the above argument, we expect its single-step prediction to be a sensible, but blurry depth map.

\PAR{Training.}
\cref{fig:overview_training} shows an overview of Marigold's training procedure.
Marigold adapts the SD v2~\cite{rombach2021stablediffusion2} UNet architecture for conditional depth map generation, using \emph{$\vect{v}$-parametrization}~\cite{salimans2022progressivedistillationfastsampling} during training.
The training objective is formulated in the latent space of the frozen SD VAE.
The GT depth map $\vect{d}^*$ is replicated along the channel dimension to conform to the 3-channel inputs of the VAE and encoded as $\vect{z}^* = \mathcal{E}(\vect{d}^*)$.
Similarly, the RGB latent is $\vect{x}=\mathcal{E}(\vect{I}_{\text{RGB}})$.
During training, noise is added only to the depth latent to get $\vect{z}_t = \sqrt{\bar\alpha_t}\vect{z}^* + \sqrt{1-\bar\alpha_t}\vect{\epsilon}$.
The UNet receives the concatenated latents and the timestep $t$ as inputs and predicts $\vect{\hat v}_t = \vect{\hat v}_\theta([\vect{z}_t, \vect{x}], t)$.
The first convolutional layer is duplicated to accomodate the larger number of input channels~\cite{ke2023marigold}.

The optimization target $\vect{v}_t^*$ at timestep $t$ is then a linear combination of the sampled noise $\vect{\epsilon}$ and the GT depth map latent $\vect{z}^*$ such that $\vect{v}_t^* = \sqrt{\bar\alpha_t}\vect{\epsilon} - \sqrt{1 - \bar\alpha_t} \vect{z}^*$.
The model is optimized with a squared error objective comparing the model prediction $\vect{\hat v}_t$ with $\vect{v}^*_t$.
With the chosen noise schedule, $t=1$ corresponds to little noise in the input, and the model is forced to predict the noise; nearer to $t=T$, the input is mostly noise and the model should predict a denoised image.
Additionally, the authors observed significantly improved depth estimation performance by training with annealed multi-resolution noise~\cite{whitaker2023pyramidnoise} and sampling with isotropic Gaussian noise~\cite{ke2023marigold}.

\PAR{Fixing Single-Step Inference.}
We now turn to analyze the behavior of Marigold during inference.
First, we show the pixel-wise standard deviation across steps for Marigold's default 50-step inference in ~\cref{fig:comparison_heatmap_std}.
We observe very small differences for almost all pixels, indicating that the model changes predictions very little during inference; in particular, this means that the first prediction of the 50-step schedule already corresponds closely to the final output.
While we would expect that the single-step output should be similar to the first step of the 50-step schedule, as shown in \cref{fig:comparison_heatmap_maribroke} it rather corresponds to pure noise.

We find that this discrepancy is caused by a flaw in the inference scheduler implementation used by Marigold and some derivative works~\cite{ke2023marigold,fu2024geowizard}.
The flaw causes the model to receive an inconsistent pairing of timestep and noise, leading to nonsensical predictions.
In particular, for a single-step prediction, the model receives a timestep encoding that indicates an almost perfect depth map whereas the actual input is pure noise.
In other words, the model receives significantly more noise than it expects, and forwards the noise almost unchanged.

Fixing the flaw is simple: we need to align the timestep with the noise level.
To do this, we can use the \texttt{\small trailing} setting as proposed in recent work~\cite{Lin2023CommonDiffusionFlawed} for image-generative models.
However, we emphasize that while this setting provided only slight improvements for image generation~\cite{Lin2023CommonDiffusionFlawed}, it is crucial for single-step inference in models such as Marigold.
In~\cref{fig:comparison_heatmap_maribroke} and~\cref{fig:comparison_heatmap_marifixed}, we compare the outputs of the same model, using the flawed and the fixed inference schedule, respectively.
Clearly, the fixed inference process produces sensible predictions, while the original does not.

%% file: sec/4_method.tex
\section{End-to-End Fine-Tuning of LDMs}

While diffusion-based depth estimation models show good overall performance and accurate details, they also exhibit artifacts such as blurred or over-sharpened outputs; see~\cref{fig:qualitative_results} for qualitative results.
This could be due to the diffusion training objective, which does not guarantee that models are trained for the desired downstream task, but for the surrogate denoising task.
To fix this, we directly fine-tune the diffusion model in an end-to-end manner.
Note that end-to-end fine-tuning of a diffusion model without sensible single-step predictions would require backpropagation through multiple network invocations, which is computationally infeasible for models with hundreds of millions of parameters.
This further shows the importance of fixing the inference pipeline as described in the previous section.

\begin{figure}
    \centering
    \includegraphics[width=\linewidth]{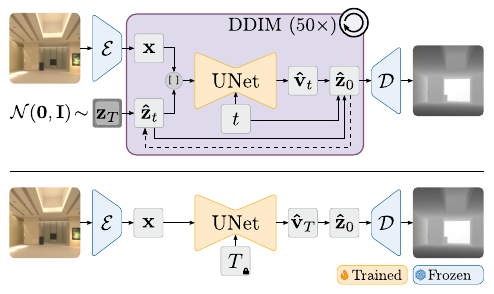}
    \caption{\textbf{Inference procedures} of Marigold (top) and our proposed simplification (bottom), which is deterministic and uses RGB latents \emph{without noise}.
    Note that the timestep is fixed to $T$ for the simplified model.
    }
    \label{fig:overview_inference}
\end{figure}

We continue to train the modified UNet used in the diffusion training stage.
However, we do not sample the timestep $t$ anymore and instead fix $t=T$ in order to always train the model for single-step prediction.
Additionally, we replace the noise with the mean of the noise distribution, \ie, zero, and only forward the RGB latent through the model.
During the diffusion training, $t=T$ corresponded to $\vect{v}_T^* = \sqrt{\bar\alpha_T}\vect{\epsilon} - \sqrt{1-\bar\alpha_T}\vect{z}^*$ according to the $\vect{v}$-parameterization~\cite{salimans2022progressivedistillationfastsampling}.
With $\bar\alpha_T\approx0$, the model is trained to convert pure noise into a clean prediction $\vect{z}^*$, effectively performing single-step prediction.
The output of the UNet can be converted into a latent depth map prediction using $\vect{\hat z}_0 = \sqrt{\bar\alpha_t} \vect{z}_t - \sqrt{1-\bar\alpha_t}\vect{\hat v}_\theta([\vect{z}_t, \vect{x}], t)$, which is decoded using the frozen VAE decoder and compared to the ground-truth depth map.
Note the difference between this fine-tuning approach and Marigold's diffusion fine-tuning objective: Marigold trains to match the \emph{latents} of the GT depth maps using an MSE loss; instead, we optimize to predict good \emph{decoded depth maps}.
Our resulting feedforward model is deterministic and we train it end-to-end using a task-specific loss.
We show a comparison of this model to the previous inference strategy in \cref{fig:overview_inference}.

For monocular depth estimation, we use an affine-invariant loss function~\cite{Ranftl2022midas} which is invariant to global scale and shift of the depth map.
In particular, we perform least-squares fitting between the ground-truth depth $\vect{d}^*$ and the predicted depth map $\vect{d}$ to estimate the scale and shift values $s$ and $t$.
The aligned prediction is then given as $\vect{\hat d} = s \vect{d} + t$, and the loss function is defined as
\begin{equation}
\label{eq:depth_loss}
    \mathcal{L}_{\text{D}} = \frac{1}{HW} \sum_{i,j} \left| d^*_{i,j} - \hat{d}_{i,j} \right|,
\end{equation}
where $(i, j)$ denotes the pixel coordinates, and $H$ and $W$ are the height and width of the image, respectively.

For surface normal estimation, we use a loss based on the angle between the ground truth and predicted normals:
\begin{equation}
\label{eq:angular_loss}
    \mathcal{L}_{\text{N}} = \frac{1}{HW} \sum_{i,j} \arccos \left( \frac{n^*_{i,j} \cdot \hat{n}_{i,j}}{\|n^*_{i,j}\| \|\hat{n}_{i,j}\|} \right),
\end{equation}
where $n^*_{i,j}$ is the ground-truth normal at pixel $(i,j)$, and $\hat{n}_{i,j}$ is the predicted normal.

%% file: sec/5_experiments.tex
\section{Experimental Setup}

\PAR{Training Datasets.}
To allow for direct comparison with Marigold~\cite{ke2023marigold}, we use the same training datasets: Hypersim~\cite{roberts2021hypersim}, which consists of photorealistic indoor scenes, and Virtual KITTI 2~\cite{cabon2020vkitti2}, which covers driving scenarios. Both datasets are fully synthetic and provide high-quality ground-truth annotations.

\begin{table*}
    \centering
    \caption{
        \textbf{Main depth estimation results.}
        Distilling Marigold with LCM is worse than simply applying the fix without any retraining.
Our task-specific training further boosts the model. We use Marigold's official evaluation pipeline to evaluate all diffusion depth models.
        Inference time is for an NVIDIA RTX 4090 GPU at 576$\times$768 resolution for a single image.
}
    \input{tbl/improve_marigold_depth}
    \label{tab:marigold_depth}
\end{table*}

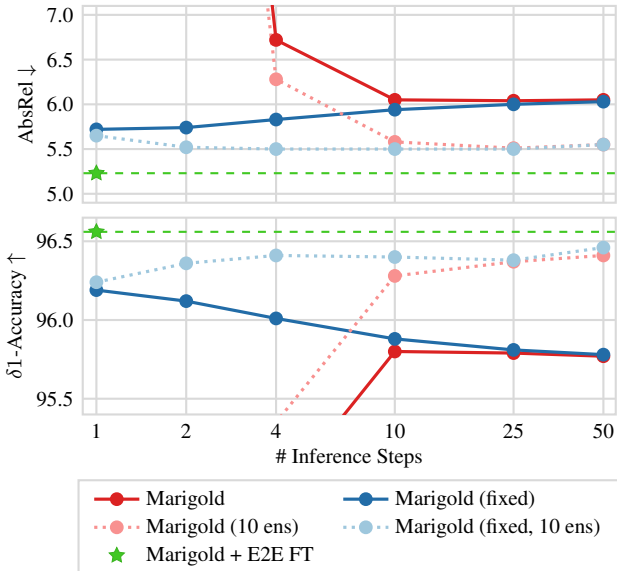
\begin{figure}
    \centering
\input{figures/comparions_nfe}\caption{
        \textbf{Depth estimation results for different numbers of inference steps.}
        The fixed inference strategy outperforms the original Marigold in all cases, but especially for a single step.
        The impact of ensembling (\textit{10 ens}) is still noticeable when performing multi-step inference.
        The deterministic end-to-end fine-tuned baseline outperforms all other versions of Marigold.}
    \label{fig:comparison_nfe}
\end{figure}

\PAR{Evaluation Datasets.}
We evaluate the fine-tuned models on commonly used benchmarks for monocular depth estimation. NYUv2~\cite{silberman2012nyuv2} and ScanNet~\cite{dai2017scannet} provide RGB-D data of indoor environments captured with Kinect cameras. ETH3D~\cite{schops2017eth3d} and DIODE~\cite{vasiljevic2019diode} consist of both indoor and outdoor scenes, derived from LiDAR sensors. KITTI~\cite{geiger2012kitti} contains outdoor driving scenes captured by vehicle-mounted cameras and LiDAR sensors.
For surface normal estimation, we evaluate on NYUv2, ScanNet, and additionally on iBims-1~\cite{koch2018ibims}, a high-quality indoor RGB-D dataset, as well as Sintel~\cite{butler2012sintel}, a synthetic outdoor dataset.

\PAR{Evaluation Protocol.}
All evaluations are conducted in the zero-shot setting.
We evaluate affine-invariant depth predictions using the standard approach, which involves the same scale and shift optimization between the predicted depth and the ground truth as in the loss computation~\cite{Ranftl2022midas}.
We report the mean absolute relative error (AbsRel), defined as the average relative difference between the ground-truth depth and the aligned predicted depth at each pixel, as well as the $\delta$1 accuracy, which is the percentage of pixels where the ratio of the aligned predicted depth to the ground truth (and its inverse) is less than 1.25.

For surface normal predictions, we report the commonly used mean angular error (Mean) between the ground-truth normal vectors and the predictions, as well as the percentage of pixels with an angular error below 11.25 degrees.

\begin{table}
    \centering
    \caption{
        \textbf{Main normal estimation results.}
        Marigold for normal estimation is trained and evaluated by us.
    }
    \input{tbl/improve_marigold_normals}
    \label{tab:marigold_normals}
\end{table}

\begin{table*}
    \centering
    \caption{
        \textbf{Fixed DDIM scheduler and end-to-end fine-tuning (E2E FT) for GeoWizard~\cite{fu2024geowizard}.} We use the official code and model weights to re-evaluate the method on all datasets.
        Inference time is for a single 576$\times$768-pixel image, evaluated on an NVIDIA RTX 4090 GPU. We obtain significant speed-ups, improving results.
        GeoWizard's original results include additional post-processing steps, such as smoothing.
    }
    \input{tbl/improve_geowizard_normals.tex}
    \label{tab:geowizard_normals}
\end{table*}

\begin{table}
    \centering
    \caption{
        \textbf{Deterministic or Probabilistic.} The effect of different types of noise for task-specific fine-tuning for depth estimation.
    }

\input{tbl/ablation_fixed_noise}
    \label{tab:ablation_fixed_noise}
\end{table}

\PAR{Implementation Details.}
For depth estimation, we use the official Marigold checkpoint, whereas for normal estimation, we train a model with the same training setup as Marigold's depth estimation, encoding normal maps as 3D vectors in the color channels.

Unless noted otherwise, we follow
Marigold's hyperparameters. 
We train all models for 20K iterations using the AdamW optimizer~\cite{loshchilov2017adamw} with a base learning rate of \num{3e-5} and an exponential learning rate decay after a 100-step warm-up.
The batch size is set to 2, with gradient accumulation over 16 steps for an effective batch size of 32.
This is a deliberate strategy to allow for mixing of images with different aspect ratios and resolutions.
We use a specific mix of indoor and outdoor scenes from both Hypersim~\cite{roberts2021hypersim} (90\%) and Virtual KITTI 2~\cite{cabon2020vkitti2} (10\%), which was beneficial to the model's performance.
Fine-tuning takes approximately 3 days on a single Nvidia H100 GPU.

\begin{figure*}
    \centering
    \newcommand{\figblock}[3]{\begin{tikzpicture}\node[anchor=south west, inner sep=0] (img1) at (0,0) {\includegraphics[width=#1]{figures/raw/colored_640/#2.png}};\begin{scope}\clip ($(img1.north west)!0.333!(img1.north east)$) -- (img1.north east) -- (img1.south east) -- ($(img1.south west)!0.667!(img1.south east)$) -- cycle;\node[anchor=south west, inner sep=0] at (0,0) {\includegraphics[width=#1]{figures/raw/normal_640/#3.png}};\end{scope}\end{tikzpicture}}\newcommand{\figline}[1]{\includegraphics[width=0.19\textwidth]{figures/raw/rgb/#1_resized_640.jpg}&\figblock{0.19\textwidth}{#1_marigold_1_1}{#1_normalgold_1_1}&\figblock{0.19\textwidth}{#1_marigold_50_10}{#1_normalgold_50_10}&\figblock{0.19\textwidth}{#1_ssigold}{#1_anggold}&\figblock{0.19\textwidth}{#1_ssiwiz}{#1_ssiwiz}}
    \setlength{\tabcolsep}{0pt}
    \footnotesize
    \begin{tabularx}{\textwidth}{YYYYY}
        \figline{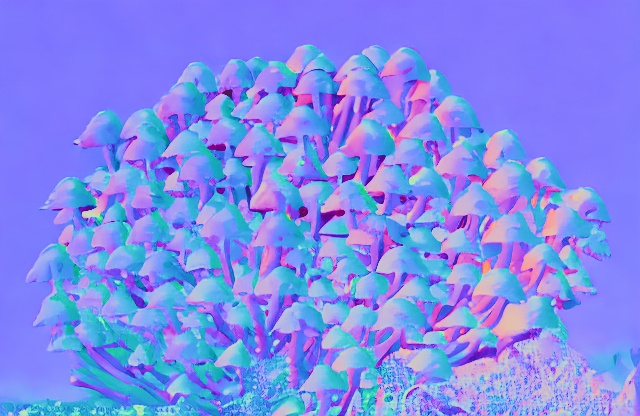}\\[-2pt]
        \figline{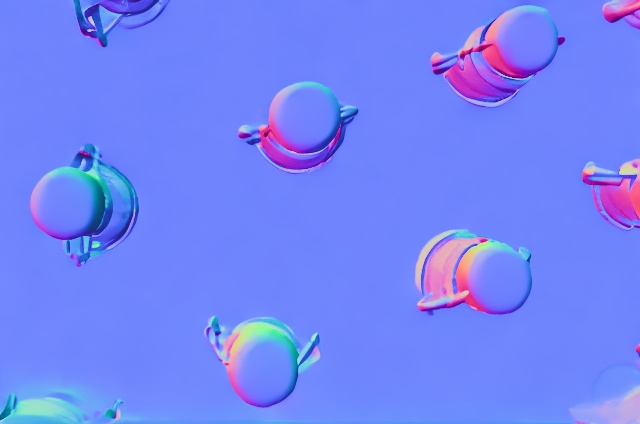}\\[0pt]
        RGB & Marigold (1, 1) & Marigold (50, 10) & Marigold + E2E FT & GeoWizard + E2E FT
    \end{tabularx}
\caption{
        \textbf{Qualitative results} of \textit{fixed} single-step/ensembled multi-step Marigold, and deterministic end-to-end fine-tuned models based on Marigold and GeoWizard.
        Multi-step results show noise-like artifacts.
        The predictions of the end-to-end fine-tuned models tend to be more sharp and accurate.
        Additional qualitative results can be found in the supplementary materials.}
    \label{fig:qualitative_results}
\end{figure*}
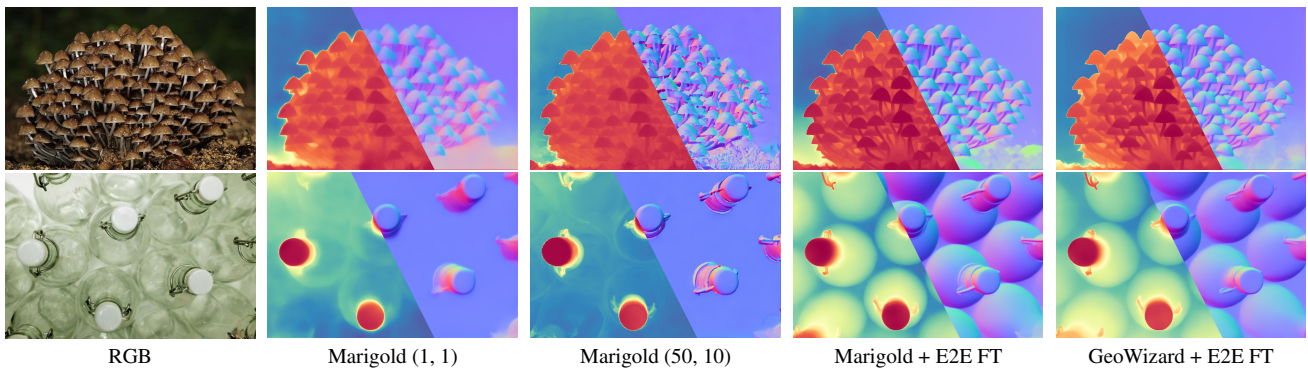

\section{Experimental Evaluation}

\PAR{Comparison with Marigold.}
As is evident from \cref{fig:comparison_nfe}, the fixed DDIM scheduler reveals that Marigold's~\cite{ke2023marigold} multi-step denoising is not actually working: instead of improving the depth map with more denoising steps, the performance actually gets \emph{worse}.
This is because repeatedly denoising sharpens the depth map, but also accumulates errors because the model expects noised ground truth latents instead of its own predictions.
This behavior was previously masked by the large error due to the broken DDIM scheduler.
Note, however, that the fixed model performs strictly better than before, for any given number of steps.
Ensembling does still provide noticeable benefits when using at least two inference steps.
For single-step inference, the predictions tend to be highly correlated, in which case ensembling does not lead to significant improvements.

We compare vanilla Marigold and a variant distilled into a Latent Consistency Model (LCM)~\cite{luo2023LCM} for few-step inference with our single-step variants in~\cref{tab:marigold_depth}.
Using 4 steps and ensemble size 5, Marigold LCM performs on par with the best setting of vanilla Marigold (50 steps, 10 ensemble) on outdoor data (KITTI~\cite{geiger2012kitti}, DIODE~\cite{vasiljevic2019diode}), but slightly worse for the indoor datasets.
Using a single step only, performance drops even more strongly to 6.5 AbsRel on NYUv2~\cite{silberman2012nyuv2}, compared to the previous best 5.5 AbsRel at 50 steps with an ensemble of 10 predictions.
In contrast, we see that vanilla Marigold with the fixed DDIM scheduler reaches 5.7 AbsRel in a single step and without ensembling.
Notably, this is better than the 6.0 AbsRel of 50-step Marigold \emph{with the same model weights}.
Moreover, we show that further end-to-end fine-tuning of Marigold leads to a substantial improvement of $-$0.5 AbsRel on NYUv2, surpassing all previous settings of Marigold, in a single step and without ensembling.
Finally, directly fine-tuning Stable Diffusion~\cite{rombach2021stablediffusion2} instead of the Marigold-pretrained model leads to comparable results.
\cref{tab:marigold_normals} paints the same overall picture for surface normal estimation; fixed single-step Marigold outperforms the vanilla multi-step, ensembled Marigold, and end-to-end fine-tuning yields even better results, even when applied to Stable Diffusion directly.

\begin{table*}
    \centering
    \caption{
        \textbf{Comparison to state-of-the-art depth estimation methods.}
        $^{\dagger}$Metric3D v2~\cite{hu2024metric3dv2} was trained on ScanNet, so zero-shot evaluation on this dataset is not possible. We gray out results that were not reproducible with the released code and models.
    }
    \input{tbl/sota_depth}
    \label{tab:sota_depth}
\end{table*}

\begin{table*}
    \centering
    \caption{
        \textbf{Comparison to state-of-the-art normal estimation methods.}
        $^{\dagger}$Metric3D v2~\cite{hu2024metric3dv2} was trained on ScanNet, so zero-shot evaluation on this dataset is not possible. We gray out results that were not reproducible with the released code and models.
}
    \input{tbl/sota_normals}
    \label{tab:sota_normals}
\end{table*}

\PAR{Deterministic or Probabilistic.}
We perform an ablation on the type of noise used during fixed-timestep fine-tuning; the results are shown in~\cref{tab:ablation_fixed_noise}.
``Gaussian" and ``Pyramid" refer to the standard normal and multi-resolution noise commonly employed in diffusion training and used in Marigold, respectively.
``Zeros" describes our default setting, \ie, no noise.
We find that using constant zeros performs slightly better than the alternatives, although the method seems to be fairly robust to the actual choice of noise.

\PAR{Comparison with GeoWizard.}
GeoWizard~\cite{fu2024geowizard} jointly predicts depth and surface normals, thus we also jointly fine-tune the model end-to-end for both tasks.
\cref{tab:geowizard_normals} shows substantial improvements for surface normal estimation.
In particular, the fine-tuned model performs substantially better than both the fixed single-step model and the claimed previous best results with 50 steps and ensembling of 10 predictions, which we were not able to reproduce. 
For depth, we observe smaller, but consistent improvements.
We provide more results in the supplementary materials.

\PAR{Qualitative Results.}
\cref{fig:qualitative_results} shows several qualitative examples. The single-step model fails to produce sharp results, while the multi-step ensemble method starts to hallucinate high frequency details. The end-to-end fine-tuned models predict sharp depth maps and high-quality normals.

\PAR{State-of-the-art Landscape.}
As shown in \cref{tab:sota_depth}, the fine-tuned models outperform current state-of-the-art generative methods for depth estimation on most datasets. Among discriminative methods, only Depth Anything~\cite{yang2024depthanything,yang2024depthanythingv2} and Metric3D~\cite{yin2023metric3d,hu2024metric3dv2} demonstrate superior performance; however, these methods were trained on datasets that are two to three orders of magnitude larger.
For surface normal estimation, the fine-tuned models set new state-of-the-art results across all evaluated datasets, with the exception of NYUv2, where Metric3D v2 continues to lead, as shown in \cref{tab:sota_normals}.

%% file: tbl/improve_marigold_depth.tex
\footnotesize
\setlength{\tabcolsep}{2.5pt}
\begin{tabularx}{\textwidth}{
lccccc
YYc
YYc
YYc
YYc
YY
}

\toprule

\mr[l]{Method} &
\mr{Steps} &
\mr{Ensemble} &&
\mr{Inference \\time} &&
\multicolumn{2}{c}{NYUv2 \cite{silberman2012nyuv2}} & &
\multicolumn{2}{c}{KITTI \cite{geiger2012kitti}} & &
\multicolumn{2}{c}{ETH3D \cite{schops2017eth3d}} & &
\multicolumn{2}{c}{ScanNet \cite{dai2017scannet}} & &
\multicolumn{2}{c}{DIODE \cite{vasiljevic2019diode}} \\

\cmidrule{7-8}\cmidrule{10-11}\cmidrule{13-14}\cmidrule{16-17}\cmidrule{19-20}

& 
& 
& 
&
&
&
AbsRel↓ & 
$\delta$1↑ & &
AbsRel↓ & 
$\delta$1↑ & &
AbsRel↓ & 
$\delta$1↑ & &
AbsRel↓ & 
$\delta$1↑ & &
AbsRel↓ & 
$\delta$1↑ \\

\midrule

Marigold \cite{ke2023marigold}
&
50 &
10 &&
\SI{24}{\second} && 5.5 & 
96.4 & &
\underline{9.9} & 
{91.6} & &
{6.5} & 
\textbf{96.0} & &
\underline{6.4} & 
{95.1} & &
{30.8} & 
{77.3} \\

Marigold \cite{ke2023marigold} 
&
50&
1&&
\SI{3.1}{\second} && {6.0} &
{95.9} & &
10.5 &
{90.4} & &
{7.1} & 
{95.1} & &
{6.9} &
{94.5} & &
31.0 &
77.2  \\

Marigold LCM &
4&
5 &&
\SI{1.8}{\second} && {6.2} &
{95.6} & &
\underline{9.9} &
91.7 & &
6.9 & 
{95.5} & &
{7.0} &
{94.5} & &
30.9 &
\underline{77.6}  \\

Marigold LCM &
1&
1&&
\textbf{\SI{121}{\milli\second}} && {6.5} &
{95.4} & &
10.7 &
89.9 & &
7.5 & 
{94.5} & &
{7.6} &
{93.8} & &
31.5 &
76.3 \\

\midrule

Marigold + DDIM fix
&
1&
1&&
\textbf{\SI{121}{\milli\second}} &&
5.7 & 
96.2 & &
10.8 & 
89.6 & &
6.9 & 
95.5 & &
6.6 & 
95.2 & &
31.1 & 
76.8 \\

Marigold + E2E FT
&
1&
1&&
\textbf{\SI{121}{\milli\second}} &&
\textbf{5.2} & \textbf{96.6} & &
\textbf{9.6} & \underline{91.9} & &
\textbf{6.2} &  \underline{95.9} & &
\textbf{5.8} & \underline{96.2} & &
\textbf{30.2} &  \textbf{77.9} \\

Stable Diffusion~\cite{rombach2021stablediffusion2} + E2E FT
&
1&
1&&
\textbf{\SI{121}{\milli\second}} &&
 \underline{5.4} & \underline{96.5} & &
\textbf{9.6} & \textbf{92.1} & &
\underline{6.4} &  \underline{95.9} & &
\textbf{5.8} & \textbf{96.5} & &
\underline{30.3} &  \underline{77.6} \\

\bottomrule

\end{tabularx}

%% file: figures/comparions_nfe.tex
\begin{tikzpicture}[tight background, font=\footnotesize]
\begin{groupplot}[
    group style={
        group size=1 by 2,
        vertical sep=0.2cm},
    xmode=log,
    xmin=0.9,
    xmax=55,
    xtick={1,2,4,10,25,50},
    xticklabels={1,2,4,10,25,50},
    legend style={at={(0.5, -0.2)}, legend columns=2,
    legend cell align={left},
    draw=none,
    inner ysep=-1.5pt,
    inner xsep=2pt,
    /tikz/every even column/.append style={column sep=0.3cm}}, width=1.04\linewidth, height=4.2cm,  grid=both, tick style={draw=none}, major grid style={line width=0.8pt,draw=gray!30}, axis line style={draw=gray!30, line width=0.8pt},
    xlabel shift={-3pt},
    ylabel shift={-3pt}
]

\nextgroupplot[
    ylabel={AbsRel ↓},
    xticklabel=\empty,
    ymin=4.9, ymax=7.1,
    ytick={5,5.5,6,6.5,7},
    yticklabels={5.0,5.5,6.0,6.5,7.0},
    legend to name=sharedlegend]
\addplot[color=marigold_col, very thick, mark=*, mark options={fill=marigold_col, solid}] table[x=steps, y= mari_absrel, col sep=comma] {data/figure_data.csv};
\addlegendentry{Marigold};

\addplot[color=marigold_fixed_col, very thick, mark=*, mark options={fill=marigold_fixed_col, solid}] table[x=steps, y= marifix_absrel, col sep=comma] {data/figure_data.csv};
\addlegendentry{Marigold (fixed)};

\addplot[color=marigold10_col, dotted, very thick, mark=*, mark options={fill=marigold10_col, solid}] table[x=steps, y= mari10_absrel, col sep=comma] {data/figure_data.csv};
\addlegendentry{Marigold (10 ens)};

\addplot[color=marigold10_fixed_col, dotted, very thick, mark=*, mark options={fill=marigold10_fixed_col, solid}] table[x=steps, y= mari10fix_absrel, col sep=comma] {data/figure_data.csv};
\addlegendentry{Marigold (fixed, 10 ens)};

\pgfplotstableread[col sep=comma]{data/figure_data.csv}\loadedtable
\pgfplotstablegetelem{0}{ours_absrel}\of\loadedtable
\let\yvalue\pgfplotsretval
\addplot [ours_col, dashed, thick, domain=0.1:60, forget plot] {\yvalue};
\addplot [mark=mystar, only marks] coordinates {(1, \yvalue)};
\addlegendentry{Marigold + E2E FT};

\nextgroupplot[
    ylabel={$\delta 1$-Accuracy ↑},
    ymin=95.4, ymax=96.65,
    ytick={95.5,96,96.5},
    yticklabels={95.5, 96.0, 96.5},
    xlabel={\# Inference Steps},]
\addplot[color=marigold_col, very thick, mark=*, mark options={fill=marigold_col, solid}] table[x=steps, y= mari_delta, col sep=comma] {data/figure_data.csv};
\addplot[color=marigold_fixed_col, very thick, mark=*, mark options={fill=marigold_fixed_col, solid}] table[x=steps, y= marifix_delta, col sep=comma] {data/figure_data.csv};
\addplot[color=marigold10_col, dotted, very thick, mark=*, mark options={fill=marigold10_col, solid}] table[x=steps, y= mari10_delta, col sep=comma] {data/figure_data.csv};
\addplot[color=marigold10_fixed_col, dotted, very thick, mark=*, mark options={fill=marigold10_fixed_col, solid}] table[x=steps, y= mari10fix_delta, col sep=comma] {data/figure_data.csv};

\pgfplotstableread[col sep=comma]{data/figure_data.csv}\loadedtable
\pgfplotstablegetelem{0}{ours_delta}\of\loadedtable
\let\yvalue\pgfplotsretval
\addplot [ours_col, dashed, thick, domain=0.1:60] {\yvalue};
\addplot [mark=mystar, only marks] coordinates {(1, \yvalue)};

\end{groupplot}

\node[draw=gray!30, line width=0.8pt, ] at ($(group c1r2.south) - (0, 1.5cm)$) {\pgfplotslegendfromname{sharedlegend}};
\end{tikzpicture}

%% file: tbl/improve_marigold_normals.tex
\scriptsize
\setlength{\tabcolsep}{1.5pt}
\begin{tabularx}{\linewidth}{
l
YYc
YYc
YYc
YY
}

\toprule

\mr[l]{Method} &
\multicolumn{2}{c}{NYUv2~\cite{silberman2012nyuv2}} & &
\multicolumn{2}{c}{ScanNet~\cite{dai2017scannet}} & &
\multicolumn{2}{c}{iBims-1~\cite{koch2018ibims}} & &
\multicolumn{2}{c}{Sintel~\cite{butler2012sintel}} \\

\cmidrule{2-3}\cmidrule{5-6}\cmidrule{8-9}\cmidrule{11-12}

&
\tiny Mean↓ & 
\tiny 11.25$^\circ$↑ & &
\tiny Mean↓ & 
\tiny 11.25$^\circ$↑ & &
\tiny Mean↓ & 
\tiny 11.25$^\circ$↑ & &
\tiny Mean↓ & 
\tiny 11.25$^\circ$↑ \\

\midrule

Marigold (50, 10)~\cite{ke2023marigold} &
18.8 &  
55.9  & & 
17.7 &  
58.8  & &  
18.4 & 
64.3  & & 
39.1 &
14.9 \\

\midrule

Marigold + DDIM fix &
17.4 & 
56.5  & &
\underline{16.8} & 
57.6  & &
18.1 & 
62.9  & &
\underline{37.1} & 
15.7 \\

Marigold + E2E FT &
\textbf{16.2} & \textbf{61.4} & &
\textbf{14.7} & \underline{66.0} & &
\textbf{15.8} & \textbf{69.9} & &
\textbf{33.5} & 
\underline{21.5} \\

SD~\cite{rombach2021stablediffusion2} + E2E FT &
\underline{16.5} & \underline{60.4} & &
\textbf{14.7} & \textbf{66.1} & &
\underline{16.1} & \underline{69.7} & &
\textbf{33.5} &
\textbf{22.3} \\

\bottomrule

\end{tabularx}

%% file: tbl/improve_geowizard_normals.tex
\footnotesize
\setlength{\tabcolsep}{2.5pt}
\begin{tabularx}{\textwidth}{
lccc
YYc
YYc
YYc
YY
}

\toprule

\mr[l]{Method} &
\mr{Steps} &
\mr{Ensemble} &
\mr{Inference \\time} & \multicolumn{2}{c}{NYUv2 \cite{silberman2012nyuv2}} & &
\multicolumn{2}{c}{ScanNet \cite{dai2017scannet}} & &
\multicolumn{2}{c}{iBims-1 \cite{koch2018ibims}} & &
\multicolumn{2}{c}{Sintel \cite{butler2012sintel}} \\

\cmidrule{5-6}\cmidrule{8-9}\cmidrule{11-12}\cmidrule{14-15}
& 
& 
& 
& 
Mean↓ & 
11.25$^\circ$↑ & &
Mean↓ & 
11.25$^\circ$↑ & &
Mean↓ & 
11.25$^\circ$↑ & &
Mean↓ & 
11.25$^\circ$↑ \\

\midrule

GeoWizard \cite{fu2024geowizard} {\scriptsize (ECCV 24)} &
50 &
10 & 
\SI{72}{\second} &
\textcolor{noreproduce}{17.0} & 
\textcolor{noreproduce}{56.5}  & &
\textcolor{noreproduce}{15.4} & 
\textcolor{noreproduce}{61.6}  & &
\textcolor{noreproduce}{13.0} & 
\textcolor{noreproduce}{65.3}  & &
--- & 
--- \\

\downrightarrow reproduced by us &
50 &
10 & 
\SI{72}{\second} &
19.1 & 
 49.5 & &
17.3 & 
53.7  & &
19.5 & 
61.6  & &
40.4 & 
13.2 \\

\midrule

GeoWizard + DDIM fix &
1 &
1 & 
\textbf{\SI{254}{\milli\second}}  &
\underline{17.0} & 
\underline{54.1}  & &
\underline{15.5} & 
\underline{59.3}  & &
\underline{18.3} & 
\underline{62.5}  & &
\underline{35.9} & 
\underline{15.6} \\

GeoWizard + E2E FT &
1 &
1 & 
\textbf{\SI{254}{\milli\second}} &
\textbf{16.1} & \textbf{60.7} & &
\textbf{15.3} & \textbf{63.6} & &
\textbf{16.2} & \textbf{69.4} & &
\textbf{33.4} & 
\textbf{22.4} \\

\bottomrule

\end{tabularx}

%% file: tbl/ablation_fixed_noise.tex
\scriptsize
\setlength{\tabcolsep}{1.5pt}
\begin{tabularx}{\linewidth}{
l
YYc
YYc
YYc
YYc
YY
}

\toprule

\mr[l]{Noise} &
\multicolumn{2}{c}{NYUv2 \cite{silberman2012nyuv2}} & &
\multicolumn{2}{c}{KITTI \cite{geiger2012kitti}} & &
\multicolumn{2}{c}{ETH3D \cite{schops2017eth3d}} & &
\multicolumn{2}{c}{ScanNet \cite{dai2017scannet}} & &
\multicolumn{2}{c}{DIODE \cite{vasiljevic2019diode}} \\

\cmidrule{2-3} 
\cmidrule{5-6} 
\cmidrule{8-9} 
\cmidrule{11-12} 
\cmidrule{14-15} 

& 
{\notsotiny AbsRel↓} & 
{\notsotiny $\delta$1↑} & &
{\notsotiny AbsRel↓} & 
{\notsotiny $\delta$1↑} & &
{\notsotiny AbsRel↓} & 
{\notsotiny $\delta$1↑} & &
{\notsotiny AbsRel↓} & 
{\notsotiny $\delta$1↑} & &
{\notsotiny AbsRel↓} & 
{\notsotiny $\delta$1↑} \\

\midrule

\multicolumn{15}{c}{Marigold~\cite{ke2023marigold} fine-tuning} \\

\midrule

Gaussian  & 
\underline{5.3} & 96.4 & &
\underline{9.9} & \underline{91.4} & &
\underline{6.3} &  \underline{95.9} & &
\underline{5.9} & \underline{96.0} & &
30.5  &  77.3 \\

Pyramid &
5.4 & \underline{96.5} & &
\underline{9.9} & 91.0 & &
\underline{6.3} &  \textbf{96.0} & &
6.0 & 95.9 & &
\textbf{30.1} &  \underline{77.7} \\

Zeros &
\textbf{5.2} & \textbf{96.6} & &
\textbf{9.6} & \textbf{91.9} & &
\textbf{6.2} &  \underline{95.9} & &
\textbf{5.8} & \textbf{96.2} & &
\underline{30.2} &  \textbf{77.9} \\

\midrule

\multicolumn{15}{c}{Stable Diffusion~\cite{rombach2021stablediffusion2} fine-tuning} \\

\midrule

Gaussian & 
5.8  & 96.1 & &
9.8 & 91.5 & &
6.6 &  95.5 & &
6.0 & 96.1 & &
30.7  &  77.2 \\

Zeros &
\textbf{5.4} & \textbf{96.5} & &
\textbf{9.6} & \textbf{92.1} & &
\textbf{6.4} &  \textbf{95.9} & &
\textbf{5.8} & \textbf{96.5} & &
\textbf{30.3} &  \textbf{77.6} \\

\bottomrule

\end{tabularx}

%% file: tbl/sota_depth.tex
\footnotesize
\setlength{\tabcolsep}{2.0pt}
\begin{tabularx}{\textwidth}{
lcc
YYc
YYc
YYc
YYc
YY
}

\toprule

\mr[l]{Method} &
\mr{{\shortstack{\text{Training}\\\text{samples}}}} &
&
\multicolumn{2}{c}{NYUv2 \cite{silberman2012nyuv2}} & &
\multicolumn{2}{c}{KITTI \cite{geiger2012kitti}} & &
\multicolumn{2}{c}{ETH3D \cite{schops2017eth3d}} & &
\multicolumn{2}{c}{ScanNet \cite{dai2017scannet}} & &
\multicolumn{2}{c}{DIODE \cite{vasiljevic2019diode}} \\

\cmidrule{4-5}\cmidrule{7-8}\cmidrule{10-11}\cmidrule{13-14}\cmidrule{16-17}

& 
&
&
AbsRel↓ & 
$\delta$1↑ & &
AbsRel↓ & 
$\delta$1↑ & &
AbsRel↓ & 
$\delta$1↑ & &
AbsRel↓ & 
$\delta$1↑ & &
AbsRel↓ & 
$\delta$1↑
\\

\midrule

MiDaS \cite{Ranftl2022midas}~\ven{TPAMI '22} &
2M &&
11.1 &
88.5 &&
23.6 &
63.0 &&
18.4 &
75.2 &&
12.1 &
84.6 &&
33.2 &
71.5
\\

LeReS \cite{yin2021leres}~\ven{CVPR '21} &
354K &&
9.0 &
91.6 &&
14.9 &
78.4 &&
17.1 &
77.7 &&
9.1 & 
91.7 &&
27.1 &
76.6
\\

Omnidata v1 \cite{eftekhar2021omnidata}~\ven{ICCV '21} &
12.2M &&
7.4 &
94.5 &&
14.9 & 
83.5 &&
16.6 &
77.8 && 
\underline{7.5} & 
93.6 &&
33.9 &
74.2
\\

HDN \cite{zhang2022hdn}~\ven{NeurIPS '22} &
300K &&
6.9 &
94.8 &&
11.5 & 
86.7 & &
12.1 &
83.3 & &
8.0 &
\underline{93.9}  &&
24.6 &
78.0
\\

DPT \cite{ranftl2021dpt}~\ven{ICCV '21} &
1.39M && 9.8 &
90.3 & &
10.0 & 
90.1 &&
7.8 &
94.6 &&
8.2 &
93.4 &&
18.2 &
75.8
\\

Depth Anything \cite{yang2024depthanything}~\ven{CVPR '24} &
62M &&
 \textbf{4.3} & 
\textbf{98.1} & &
 7.6 & 
94.7 & &
12.7 & 
88.2 & &
--- &
--- & &
\underline{6.6} &
\underline{95.2}
\\

Depth Anything v2 \cite{yang2024depthanythingv2}~\ven{arXiv '24} &
62M &&
\underline{4.4} & 
\underline{97.9} & &
7.5 & 
94.8 & &
13.2 &
86.2 & &
--- &
--- & &
\textbf{6.5} &
\textbf{95.4}
\\

Metric3D \cite{yin2023metric3d}~\ven{ICCV '23} &
8M &&
5.0 &
96.6 & &
\underline{5.8} &
\underline{97.0} & &
\underline{6.4} &
\underline{96.5} & &
\textbf{7.4} &
\textbf{94.1} & &
22.4 &
80.5
\\

Metric3D v2 \cite{hu2024metric3dv2}~\ven{TPAMI '24} &
16M &&
\textbf{4.3} &
\textbf{98.1} & &
\textbf{4.4} &
\textbf{98.2} & &
\textbf{4.2} &
\textbf{98.3} & &
---\textsuperscript{\textdagger} &
---\textsuperscript{\textdagger} & &
13.6 &
89.5
\\

\midrule

Marigold \cite{ke2023marigold}~\ven{CVPR '24} &
74K &&
5.5 & 
96.4 & &
\underline{9.9} & 
91.6 & &
6.5 & 
\underline{96.0} & &
{6.4} & 
{95.1} & &
30.8 & 
{77.3} \\

GeoWizard \cite{fu2024geowizard}~\ven{ECCV '24} &
278K && \textcolor{noreproduce}{5.2} & 
\textcolor{noreproduce}{96.6} & &
\textcolor{noreproduce}{9.7} & 
\textcolor{noreproduce}{92.1} & &
\textcolor{noreproduce}{6.4} & 
\textcolor{noreproduce}{96.1} & &
\textcolor{noreproduce}{6.1} & 
\textcolor{noreproduce}{95.3} & &
\textcolor{noreproduce}{29.7} & 
\textcolor{noreproduce}{79.2} \\

\downrightarrow reproduced by us & 
278K && {5.7} & 
{96.2} & &
{14.4} &
{82.0} & &
{7.5} & 
{94.3} & &
\underline{6.1} & 
95.8 & &
{31.4} & 
{77.1} \\

DepthFM \cite{gui2024depthfm}~\ven{arXiv '24} &
63K &&
\textcolor{noreproduce}{6.5}  & 
\textcolor{noreproduce}{95.6} & &
\textcolor{noreproduce}{8.3}   & 
\textcolor{noreproduce}{93.4}  & &
\textcolor{noreproduce}{---}  & 
\textcolor{noreproduce}{---}    & &
\textcolor{noreproduce}{---}    & 
\textcolor{noreproduce}{---}    & &
\textcolor{noreproduce}{22.5}  & 
\textcolor{noreproduce}{80.0} \\

\downrightarrow reproduced by us & 63K &&
{6.9} &       
{95.4} & &   
{11.4} &     
{88.1} & &   
6.5 & 
\textbf{96.2} & &
{8.1} &    
{92.5} & & 
\textbf{25.0} &   
\textbf{78.3} \\   

\arrayrulecolor{gray!50!white}
\midrule
\arrayrulecolor{black}

Marigold + E2E FT &
74K &&
\textbf{5.2} & \textbf{96.6} & &
\textbf{9.6} & \underline{91.9} & &
\textbf{6.2} &  95.9 & &
\textbf{5.8} & \underline{96.2} & &
\underline{30.2} &  \underline{77.9} \\

Stable Diffusion \cite{rombach2021stablediffusion2} + E2E FT &
74K &&
\underline{5.4} & \underline{96.5} & &
\textbf{9.6} & \textbf{92.1} & &
\underline{6.4} &  95.9 & &
\textbf{5.8} & \textbf{96.5} & &
30.3 &  77.6 \\

\bottomrule

\end{tabularx}

%% file: tbl/sota_normals.tex
\footnotesize
\setlength{\tabcolsep}{2.0pt}
\begin{tabularx}{\textwidth}{
lcc
YYc
YYc
YYc
YY
}

\toprule
\mr[l]{Method} &
\mr{Training\\samples} & &
\multicolumn{2}{c}{NYUv2 \cite{silberman2012nyuv2}} & &
\multicolumn{2}{c}{ScanNet \cite{dai2017scannet}} & &
\multicolumn{2}{c}{iBims-1 \cite{koch2018ibims}} & &
\multicolumn{2}{c}{Sintel \cite{butler2012sintel}} \\

\cmidrule{4-5}\cmidrule{7-8}\cmidrule{10-11}\cmidrule{13-14}

& 
& &
Mean↓ & 
11.25$^\circ$↑ & &
Mean↓ & 
11.25$^\circ$↑ & &
Mean↓ & 
11.25$^\circ$↑ & &
Mean↓ & 
11.25$^\circ$↑
\\

\midrule

Omnidata v1 \cite{eftekhar2021omnidata}~\ven{ICCV '21} &
12.2M & &
23.1 & 
45.8 & & 
\underline{22.9} & 
47.4 & &
19.0 & 
62.1 & &
41.5 &  
11.4 \\

Omnidata v2 \cite{kar2022omnidatav2}~\ven{CVPR '22} &
12.2M & &
17.2 & 
55.5 & & 
\textbf{16.2} & 
\underline{60.2} & &
\underline{18.2} & 
63.9 & &
\underline{40.5} &  
\underline{14.7} \\

DSINE \cite{bae2024dsine}~\ven{CVPR '24} &
161K & &
\underline{16.4} & 
\underline{59.6}  & &
\textbf{16.2} & 
\textbf{61.0}  & &
\textbf{17.1} & 
\underline{67.4} & &
\textbf{34.9} & 
\textbf{21.5} \\

Metric3D v2 \cite{hu2024metric3dv2}~\ven{TPAMI '24} &
16M & &
\textbf{13.3} & 
\textbf{66.4}  & &
---\textsuperscript{\textdagger}  & 
---\textsuperscript{\textdagger}  & &
19.6 & 
\textbf{69.7} & &
--- & 
--- \\

\midrule

Marigold~\cite{ke2023marigold}~\ven{CVPR '24} &
74K & &
18.8 &  
55.9  & & 
17.7 &  
58.8  & &  
18.4 & 
64.3  & & 
39.1 &
14.9 \\

GeoWizard \cite{fu2024geowizard}~\ven{ECCV '24} &
278K & &
\textcolor{noreproduce}{17.0} & 
\textcolor{noreproduce}{56.5}  & &
\textcolor{noreproduce}{15.4} & 
\textcolor{noreproduce}{61.6}  & &
\textcolor{noreproduce}{13.0} & 
\textcolor{noreproduce}{65.3}  & &
\textcolor{noreproduce}{---} & 
\textcolor{noreproduce}{---} \\

\downrightarrow reproduced by us &
278K & &
19.1 & 
49.5 & &
17.3 & 
53.7  & &
19.5 & 
61.6  & &
40.4 & 
13.2 \\

\arrayrulecolor{gray!50!white}
\midrule
\arrayrulecolor{black}

GeoWizard + E2E FT &
278K & &
\textbf{16.1} & \underline{60.7} & &
\underline{15.3} & {63.6} & &
16.2 & {69.4} & &
\textbf{33.4} & 
\textbf{22.4} \\

Marigold + E2E FT &
74K & &
\underline{16.2} & \textbf{61.4} & &
\textbf{14.7} & \underline{66.0} & &
\textbf{15.8} & \textbf{69.9} & &
\underline{33.5} & 
21.5 \\

Stable Diffusion \cite{rombach2021stablediffusion2} + E2E FT &
74K & &
16.5 & 60.4 & &
\textbf{14.7} & \textbf{66.1} & &
\underline{16.1} & \underline{69.7} & &
\underline{33.5} &
\underline{22.3} \\

\bottomrule

\end{tabularx}

%% file: sec/6_discussion.tex
\section{Conclusion}
\vspace{-0.5em}
We have shown that a critical flaw in the implementation of the DDIM scheduler causes several prior works to draw possibly wrong conclusions.
We found simple end-to-end fine-tuning to outperform more complicated training pipelines and architectures.
Nonetheless, our work supports the hypothesis that diffusion pretraining does provide excellent priors for geometric tasks such as monocular depth and normal estimation.
The resulting models allow accurate single-step inference, enabling to profit from large-scale data using sophisticated self-training procedures as used in prior works~\cite{yang2024depthanything,yang2024depthanythingv2}.
We believe that further improvements in diffusion models will lead to even more reliable priors, which might further improve the performance of this kind of geometry estimation models.
We regard this as a promising avenue for future research.

{\footnotesize
\textbf{Acknowledgements.}
Karim Abou Zeid's research is funded by the Bosch-RWTH LHC project \enquote{Context Understanding for Autonomous Systems}.
Christian Schmidt is funded by BMBF project bridgingAI (16DHBKI023).
Computations were performed with computing resources granted by RWTH Aachen University under project rwth1690.
}

%% file: sec/X_suppl.tex
\appendix

\setcounter{table}{0}
\setcounter{figure}{0}

\renewcommand{\thetable}{A-\arabic{table}}

\renewcommand{\thefigure}{A-\arabic{figure}}

\twocolumn[
    \begin{center}
        \Large\bf Fine-Tuning Image-Conditional Diffusion Models is Easier than You Think \par
        \vspace*{12pt}
        \large Supplementary Material
        \vspace*{12pt}
    \end{center}
]

\newcommand{\sdprompt}{Cosy small dark living room, no people}

\section{DDIM Inference}
\label{sec:supp:ddim_inference}
During training, the highest noise level corresponds to the last timestep $t = T$, and $t=1$ corresponds to a very small noise level.
The DDIM inference scheduler iterates over a series of $k$ timesteps $\tau_1 > \tau_2 > \ldots > \tau_k > 0$ and iteratively denoises the initial noise input $\vect{z}_{\tau_1}$.
We consider the \texttt{leading} and \texttt{trailing} schedules that are also discussed by Lin~\etal~\cite{Lin2023CommonDiffusionFlawed} and show the selected timesteps for different $k$ in~\cref{tab:leading_vs_trailing}.
The original \texttt{leading} timestep selection strategy of the DDIM scheduler excludes the final timestep $T$.
This leads to a mismatch between training and inference; using the \texttt{leading} schedule, the model receives noise as input, even though the timestep embedding indicates a partially denoised input.
In contrast, the fixed \texttt{trailing} strategy always starts with $t=T$ for the first denoising step, properly aligning training and inference.
In the limit of $k\rightarrow T$ inference steps, both strategies converge to the same behavior.

\begin{table*}[!htbp]
\centering
\footnotesize
\caption{
\textbf{Comparison of \texttt{leading} vs. \texttt{trailing} timestep selection.}
The timesteps selected by two DDIM scheduler timestep selection strategies for $T=1000$ timesteps and varying numbers of inference steps.
}
\begin{tabularx}{\textwidth}{cXX}
\hline
\toprule
Inference Steps & \texttt{leading} timestep selection & \texttt{trailing} timestep selection\\
\midrule
1 & {[}1{]} & {[}1000{]} \\
2 & {[}501, 1{]} & {[}1000, 500{]} \\
4 & {[}751, 501, 251, 1{]} & {[}1000, 750, 500, 250{]} \\
10 & {[}901, 801, 701, 601, 501, 401, 301, 201, 101, 1{]} & {[}1000, 900, 800, 700, 600, 500, 400, 300, 200, 100{]} \\
\bottomrule
\label{tab:leading_vs_trailing}
\end{tabularx}
\end{table*}

\begin{figure}
    \centering
    \begin{adjustbox}{max width=\linewidth}
    \begin{tikzpicture}[
        image/.style={
            inner sep=0pt,
            outer sep=1pt,
        },
    ]
    \newcommand{\boxWidth}{0.33\linewidth}
    \newcommand{\mgold}{Marigold_samples_color}
    \newcommand{\sd}{SD_Full}
    \newcommand{\ddimSample}[4]{\includegraphics[width=\boxWidth]{figures/supp/ddim_samples/#1/#2_#3/#2_#3_#4}}

    \node[image] (rgb) at (0,0) {
        \includegraphics[width=\boxWidth]{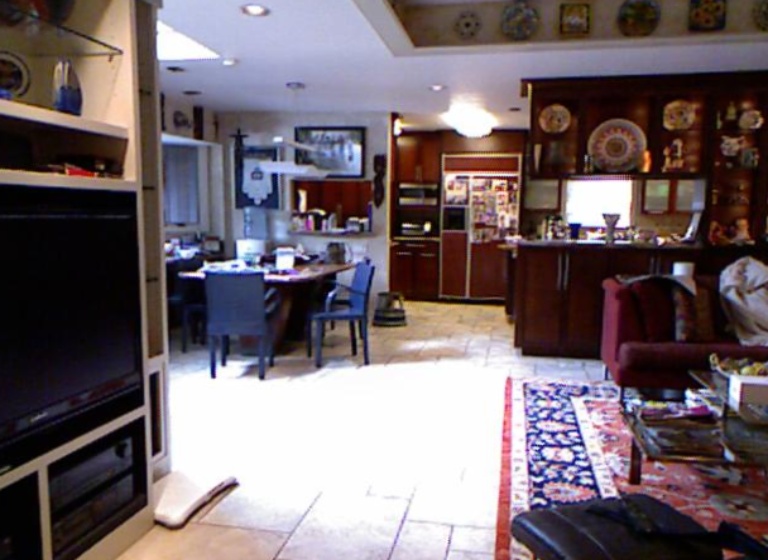}
    };
    \node [image, minimum height=\boxWidth, text width=\boxWidth, align=center, anchor=north]
        (prompt) at (rgb.south) {
        \texttt{\sdprompt}
    };

    \node [rotate=90, anchor=south] (mgold_label) at (rgb.west) {Marigold};
    \node [rotate=90, anchor=south] (sd_label) at (prompt.west) {Stable Diffusion};

    \node[image, anchor=west] (mgold_leading) at (rgb.east) {
        \ddimSample{\mgold}{leading}{1}{0}
    };
    \node[image, anchor=west] (mgold_trailing) at (mgold_leading.east) {
        \ddimSample{\mgold}{trailing}{1}{0}
    };

    \node[image, anchor=north] (sd_leading) at (mgold_leading.south) {
        \ddimSample{\sd}{leading}{1}{0}
    };
    \node[image, anchor=north] (sd_trailing) at (mgold_trailing.south) {
        \ddimSample{\sd}{trailing}{1}{0}
    };

    \node [anchor=north] (prompt_label) at (prompt.south) {Input prompt};
    \node [anchor=north] (sd_leading) at (sd_leading.south) {Broken DDIM};
    \node [anchor=north] (sd_trailing) at (sd_trailing.south) {Fixed DDIM};
    \end{tikzpicture}
    \end{adjustbox}
    \caption{
        \textbf{Single-step outputs of Marigold and Stable Diffusion.}
        With a single step, Stable Diffusion produces a blurry image at best, while Marigold outputs a sensible depth map.
        Note that the input prompt is text for Stable Diffusion, but an RGB image for Marigold.
    }
    \label{fig:comparison_singlestep_ddim}
\end{figure}

In~\cref{fig:comparison_singlestep_ddim}, we illustrate the difference between single-step predictions using the broken \texttt{leading} and the fixed \texttt{trailing} DDIM scheduler for Marigold~\cite{ke2023marigold} and Stable Diffusion~\cite{rombach2021stablediffusion2}.
Both models output noise when using the broken scheduler.
With the fixed implementation, both models predict the mean of their respective conditional distribution.
For single-step Marigold this results in a well-defined depth map, whereas for single-step Stable Diffusion, it produces a blurry image with coarse structures that roughly align with the input prompt.

\cref{fig:comparison_fewstep_ddim} further demonstrates the scheduler's impact when multiple steps are considered.
It clearly shows that the effect of the broken scheduler becomes less noticeable as the number of inference steps increases.
Additionally, the weak text conditioning in Stable Diffusion leads to blurry images, which gradually sharpen as more inference steps are taken.
In contrast, the strong image conditioning in Marigold allows the model to predict reasonably accurate depth maps already in the first step.
As shown by the heatmap in Fig. 2b in the main text, subsequent steps only lead to small changes in the predicted distances, and most of the scene remains unchanged.

\section{Detailed Experimental Setup}
\label{sec:supp:impl_details}

\paragraph{Training Datasets.}
For a direct comparison with Marigold~\cite{ke2023marigold}, we use the same synthetic training datasets offering high quality ground-truth annotations, \ie, Hypersim~\cite{roberts2021hypersim} and Virtual KITTI 2~\cite{cabon2020vkitti2}.

\begin{figure}[t]
    \centering
    \begin{minipage}{\linewidth}
        \centering
        \includegraphics[width=\linewidth]{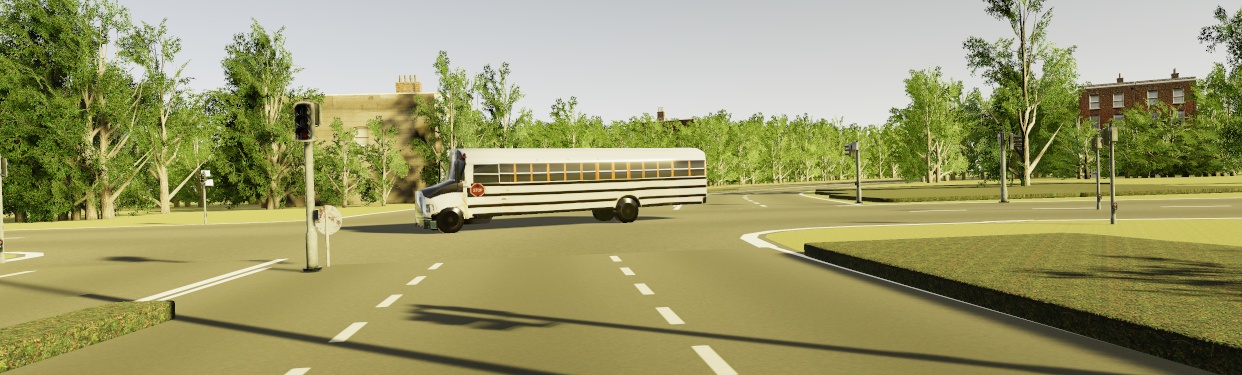}
    \end{minipage}
\begin{minipage}{\linewidth}
        \centering
        \includegraphics[width=\linewidth]{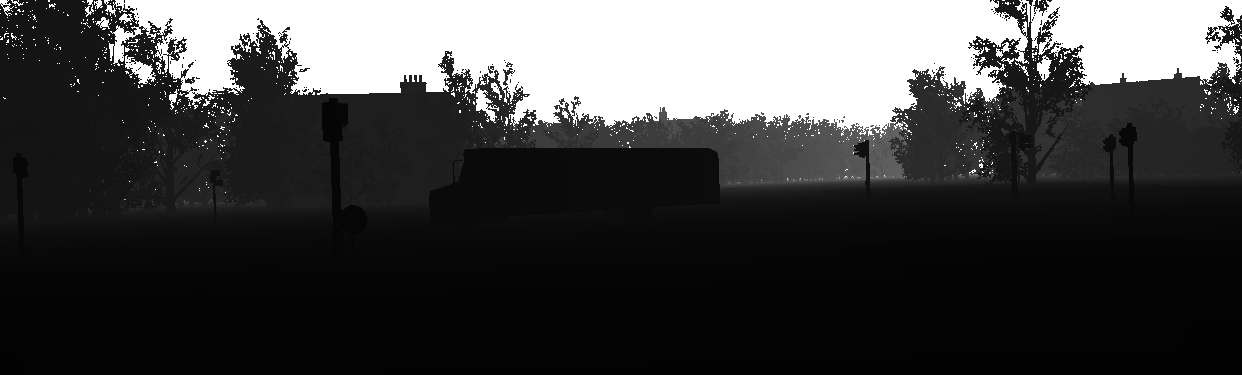}
    \end{minipage}
\begin{minipage}{\linewidth}
        \centering
        \includegraphics[width=\linewidth]{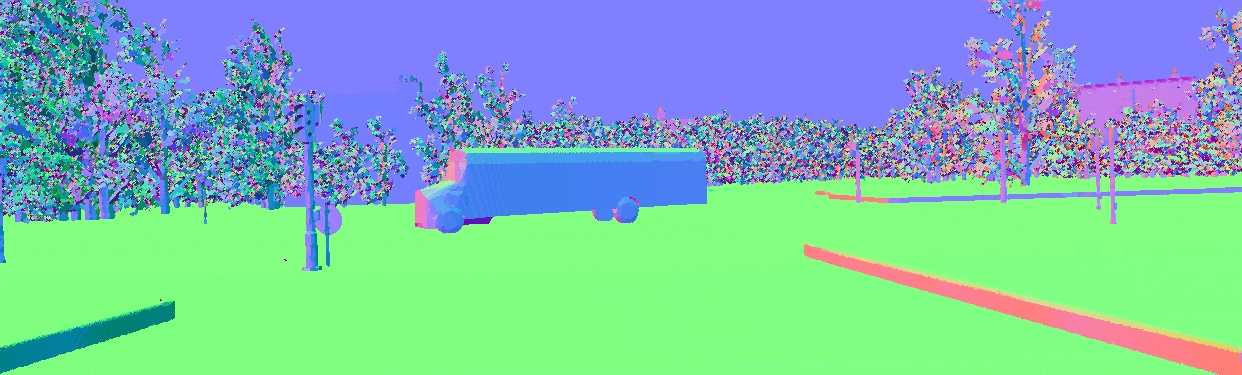}
    \end{minipage}
    \caption{\textbf{Virtual KITTI 2 example.} Top: Synthetic RGB image. Middle: Ground-truth depth map. Bottom: Ground-truth surface normals, generated using discontinuity-aware gradient filters~\cite{feng2023d2nt}.}
    \label{fig:kitti_stacked}
\end{figure}

Hypersim consists of 54K photorealistic images from 365 indoor scenes, which we resize to a resolution of $480 \times 640$ with a far plane at 65 meters. Virtual KITTI 2 contains approximately 20K samples from four synthetic driving scenarios under various weather conditions. These images are cropped to $352 \times 1216$ pixels, and the far plane is set to 80 meters. 

Since Virtual KITTI 2 does not provide annotations for surface normals, we compute them ourselves with the ground-truth depth maps, employing discontinuity-aware gradient filters from \cite{feng2023d2nt}. A qualitative example of the resulting normals can be seen in \cref{fig:kitti_stacked}.

\paragraph{Data Preprocessing.}
Following Marigold's approach for depth estimation, we remove outliers, \ie, values below the 2\textsuperscript{nd} percentile and above the 98\textsuperscript{th} percentile, and normalize the depth map to the range $[-1,1]$. Then, we repeat the normalized depth map 3 times along the color channel to match the VAE encoder's expected input shape. Normals, on the other hand, can be encoded directly since they are already in the desired range of $[-1,1]$ and match the number of channels. The only data augmentation we utilize is random horizontal flipping.

\paragraph{Training Details.}
We mask out undefined depth values in the Hypersim dataset, and pixels surpassing the far plane for Virtual KITTI 2. When training Marigold for normal prediction as a diffusion estimator, the mask is downsampled by a factor of 8 to match the latent resolution. Thus, we neither enforce nor supervise undefined regions. 
For the end-to-end fine-tuning of GeoWizard, both the scale and shift invariant depth loss and the angular loss are optimized jointly. Scaling the depth loss by a factor of 0.5 roughly ensures equal magnitude.

\paragraph{Evaluation Datasets.}
For monocular depth estimation, we follow the evaluation strategy of Marigold and evaluate on commonly used benchmarks. NYUv2~\cite{silberman2012nyuv2} and ScanNet~\cite{dai2017scannet} provide RGB-D data of indoor environments captured with Kinect cameras. We use the official NYUv2 test split, consisting of 654 instances, while for ScanNet, Marigold's set of 800 randomly sampled images from the 312 validation scenes~\cite{ke2023marigold} is employed.
ETH3D~\cite{schops2017eth3d} and DIODE~\cite{vasiljevic2019diode} offer high-resolution depth data for both indoor and outdoor scenes, derived from LiDAR sensors. We evaluate on all 454 samples in ETH3D and on DIODE's validation set, comprising 325 indoor and 446 outdoor examples.
For KITTI~\cite{geiger2012kitti}, consisting of outdoor driving scenes captured by vehicle-mounted cameras and LiDAR sensors, the Eigen test split~\cite{eigen2014depthmappredfromsingleimage} is used, containing 652 images.

Regarding surface normal estimation we utilize the official DSINE \cite{bae2024dsine} evaluation pipeline and data, comprised of the NYUv2 test split, 300 ScanNet \cite{silberman2012nyuv2} samples, the full iBims-1 \cite{koch2018ibims} dataset, which is a small high-quality RGB-D dataset of 100 samples, and Sintel \cite{butler2012sintel}, made up of 1064 synthetic outdoor examples derived from an open-source 3D animated short film.

\paragraph{Evaluation Details.}
For most existing methods in Tab. 5 and Tab. 6 we obtain the performance metrics either from the papers introducing these methods or from the Marigold and DSINE papers.
The missing scores, like those of the newer GeoWizard~\cite{fu2024geowizard} and DepthFM~\cite{gui2024depthfm} models, are obtained by reevaluating the respective models with their official inference code and released checkpoints. In the case of DepthFM, the prediction alignment with respect to the ground-truth metric depth happens in the log metric space.

\begin{table}[t]
    \centering
    \caption{
        \textbf{Frozen vs.\ fine-tuned VAE decoder.} We conduct end-to-end fine-tuning of Marigold~\cite{ke2023marigold} for depth estimation, and assess the effect of freezing or fine-tuning the weights of the pretrained VAE decoder.
    }

\input{tbl/decoder_freezing}
    \label{tab:decoder_freezing}
\end{table}

\begin{table*}[!t]
    \centering
    \caption{
        \textbf{Fixed DDIM scheduler and end-to-end fine-tuning (E2E FT) for GeoWizard's~\cite{fu2024geowizard} depth estimation.} We use the official code and model weights to re-evaluate the method on all datasets. Inference time is for a single 576$\times$768-pixel image, evaluated on an NVIDIA RTX 4090 GPU. We obtain significant speed-ups, improving results.
    }
    \input{tbl/improve_geowizard_depth.tex}
    \label{tab:geowizard_depth}
\end{table*}

\begin{table*}[!t]
    \centering
    \caption{
        \textbf{Comparison of DepthFM~\cite{gui2024depthfm} with the DDIM-fixed and end-to-end fine-tuned (E2E FT) Marigold and Stable Diffusion models.}
        We re-evaluated DepthFM~\cite{gui2024depthfm} on all datasets using the official code and model weights, with 4 inference steps and an ensemble size of 6.
        Inference time is for a single 576$\times$768-pixel image, evaluated on an NVIDIA RTX 4090 GPU.
    }
    \input{tbl/compare_depth_fm}
    \label{tab:compare_depth_fm}
\end{table*}

\begin{figure*}[p]
    \centering
    \begin{adjustbox}{max width=\textwidth, totalheight=\textheight-1.5cm}
    \begin{tikzpicture}[
        image/.style={
            inner sep=0pt,
            outer sep=1pt,
        },
    ]
    \newcommand{\boxWidth}{0.3\linewidth}
    \newcommand{\mgold}{Marigold_samples_color}
    \newcommand{\sd}{SD_Full}
    \newcommand{\ddimSample}[4]{\includegraphics[width=\boxWidth]{figures/supp/ddim_samples/#1/#2_#3/#2_#3_#4}}

    \node[image, anchor=north west] (mgold_leading20) at (0, 0) {
        \ddimSample{\mgold}{leading}{2}{0}
    };
    \node[image, anchor=west] (mgold_leading21) at (mgold_leading20.east) {
        \ddimSample{\mgold}{leading}{2}{1}
    };
    \node[rotate=90, anchor=south] at (mgold_leading20.west) {Broken DDIM, $N=2$};

    \node[image, anchor=north] (mgold_leading40) at (mgold_leading20.south) {
        \ddimSample{\mgold}{leading}{4}{0}
    };
    \foreach \i [evaluate=\i as \ii using int(\i-1)] in {1,2,3}{
    \node[image, anchor=west] (mgold_leading4\i) at (mgold_leading4\ii.east) {
        \ddimSample{\mgold}{leading}{4}{\i}
    };
    }
    \node[rotate=90, anchor=south] at (mgold_leading40.west) {Broken DDIM, $N=4$};

    \node[image, anchor=north west] (mgold_trailing20) at (mgold_leading40.south west) {
        \ddimSample{\mgold}{trailing}{2}{0}
    };
    \node[image, anchor=west] (mgold_trailing21) at (mgold_trailing20.east) {
        \ddimSample{\mgold}{trailing}{2}{1}
    };
    \node[rotate=90, anchor=south] at (mgold_trailing20.west) {Fixed DDIM, $N=2$};

    \node[image, anchor=north] (mgold_trailing40) at (mgold_trailing20.south) {
        \ddimSample{\mgold}{trailing}{4}{0}
    };
    \foreach \i [evaluate=\i as \ii using int(\i-1)] in {1,2,3}{
    \node[image, anchor=west] (mgold_trailing4\i) at (mgold_trailing4\ii.east) {
        \ddimSample{\mgold}{trailing}{4}{\i}
    };
    }
    \node[rotate=90, anchor=south] at (mgold_trailing40.west) {Fixed DDIM, $N=4$};

    \coordinate (mgold_center) at ($(mgold_leading20.north west)!0.5!(mgold_trailing40.south west)$);
    \node[image, anchor=east] (rgb) at ($(mgold_center.west)-(1cm,0)$) {
        \includegraphics[width=\boxWidth]{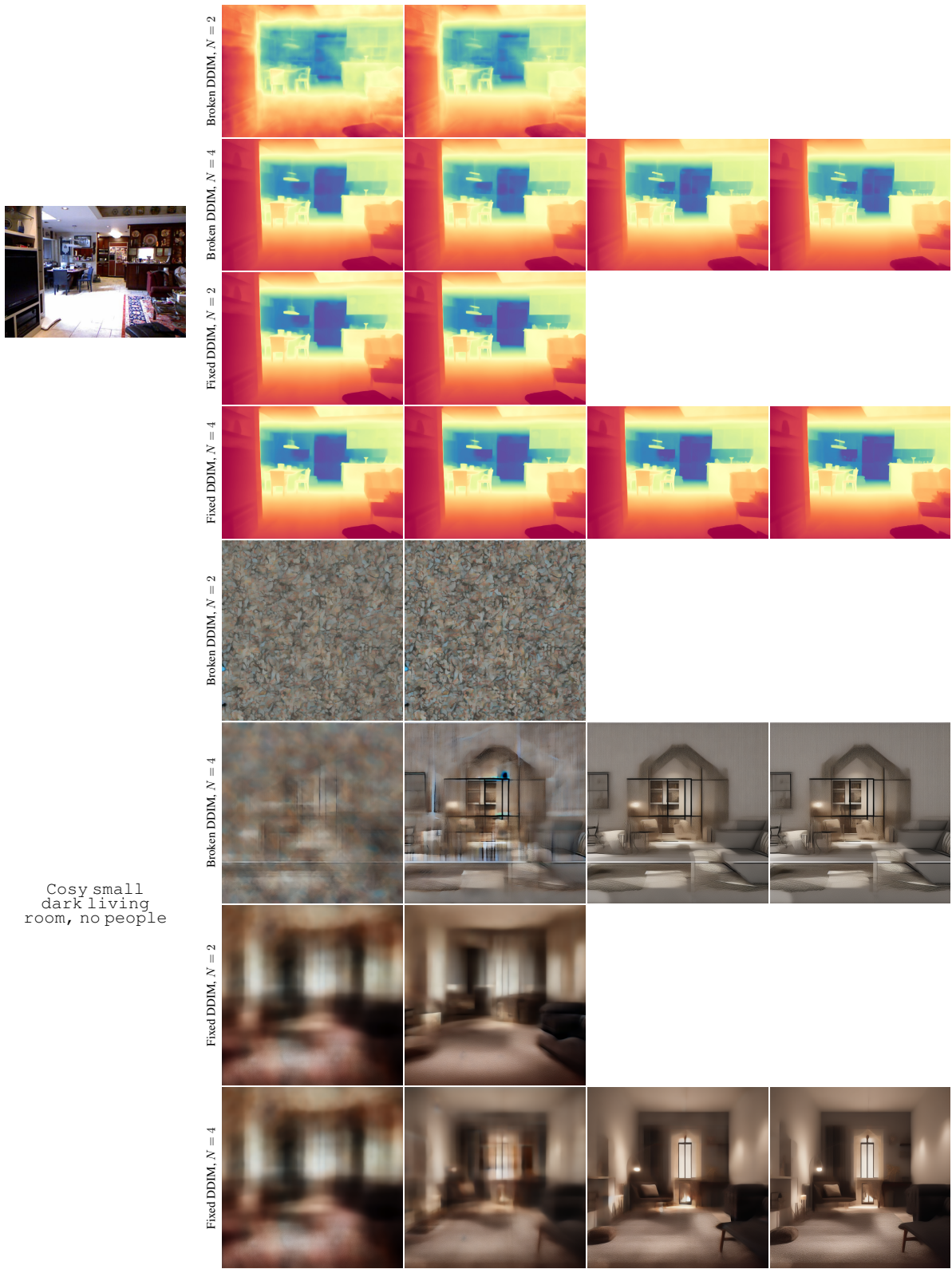}
    };

    \node[image, anchor=north west] (sd_leading20) at (mgold_trailing40.south west) {
        \ddimSample{\sd}{leading}{2}{0}
    };
    \node[image, anchor=west] (sd_leading21) at (sd_leading20.east) {
        \ddimSample{\sd}{leading}{2}{1}
    };
    \node[rotate=90, anchor=south] at (sd_leading20.west) {Broken DDIM, $N=2$};

    \node[image, anchor=north] (sd_leading40) at (sd_leading20.south) {
        \ddimSample{\sd}{leading}{4}{0}
    };
    \foreach \i [evaluate=\i as \ii using int(\i-1)] in {1,2,3}{
    \node[image, anchor=west] (sd_leading4\i) at (sd_leading4\ii.east) {
        \ddimSample{\sd}{leading}{4}{\i}
    };
    }
    \node[rotate=90, anchor=south] at (sd_leading40.west) {Broken DDIM, $N=4$};

    \node[image, anchor=north west] (sd_trailing20) at (sd_leading40.south west) {
        \ddimSample{\sd}{trailing}{2}{0}
    };
    \node[image, anchor=west] (sd_trailing21) at (sd_trailing20.east) {
        \ddimSample{\sd}{trailing}{2}{1}
    };
    \node[rotate=90, anchor=south] at (sd_trailing20.west) {Fixed DDIM, $N=2$};

    \node[image, anchor=north] (sd_trailing40) at (sd_trailing20.south) {
        \ddimSample{\sd}{trailing}{4}{0}
    };
    \foreach \i [evaluate=\i as \ii using int(\i-1)] in {1,2,3}{
    \node[image, anchor=west] (sd_trailing4\i) at (sd_trailing4\ii.east) {
        \ddimSample{\sd}{trailing}{4}{\i}
    };
    }
    \node[rotate=90, anchor=south] at (sd_trailing40.west) {Fixed DDIM, $N=4$};

    \coordinate (sd_center) at ($(sd_leading20.north west)!0.5!(sd_trailing40.south west)$);
    \node[image, text width=\boxWidth, align=center, anchor=east] (prompt) at ($(sd_center.west)-(1cm,0)$) {
        \Large \texttt{\sdprompt}
    };

    \end{tikzpicture}
    \end{adjustbox}
    \caption{
        \textbf{Few-step inference of Marigold and Stable Diffusion.}
        With more steps, the adverse effects of the broken DDIM scheduler get less pronounced.
        Both Marigold and Stable Diffusion produce sharper outputs with more steps, but the difference is much greater for Stable Diffusion.
    }
    \label{fig:comparison_fewstep_ddim}
\end{figure*}

\section{Additional Results}
\label{sec:supp:exp_results}

\paragraph{GeoWizard for Depth Estimation.}
GeoWizard~\cite{fu2024geowizard} jointly predicts depth and surface normals, using a similar training and evaluation setup as Marigold.
We find that GeoWizard suffers from the same flaw in the DDIM implementation as Marigold, and end-to-end fine-tuning the model for depth and normal estimation significantly boosts the performance (see~\cref{tab:geowizard_depth} and Tab. 3 in the main text).
In particular, the fine-tuned model performs better than both the fixed single-step model and the previously best reported results with 50 steps and ensembling of 10 predictions.

\paragraph{Further Comparisons to DepthFM.}
DepthFM~\cite{gui2024depthfm} proposes a direct mapping from input images to depth maps through flow matching, leveraging Stable Diffusion v2~\cite{rombach2021stablediffusion2} as a prior. We observe that, apart from the ETH3D $\delta$1 and DIODE~\cite{vasiljevic2019diode} metrics, a simpler approach like E2E FT achieves better performance with a more than $10\times$ speedup as seen in \cref{tab:compare_depth_fm}. 

\paragraph{Fine-Tuning the VAE Decoder.}
By default, we keep the pretrained VAE decoder frozen while conducting end-to-end fine-tuning. \cref{tab:decoder_freezing} shows that fine-tuning the weights of this decoder does not improve performance.

\begin{figure*}
    \centering
    \newcommand{\figblock}[2]{\begin{tikzpicture}\node[anchor=south west, inner sep=0] (img1) at (0,0) {\includegraphics[width=#1]{figures/raw/colored_640/#2.png}};\end{tikzpicture}}\newcommand{\figline}[1]{\includegraphics[width=0.19\textwidth]{figures/raw/rgb/#1_resized_640}&\figblock{0.19\textwidth}{#1_marigold_1_1}&\figblock{0.19\textwidth}{#1_marigold_50_10}&\figblock{0.19\textwidth}{#1_ssigold}&\figblock{0.19\textwidth}{#1_ssiwiz}}
    \setlength{\tabcolsep}{0pt}
    \footnotesize
    \begin{tabularx}{\textwidth}{YYYYY}
        \figline{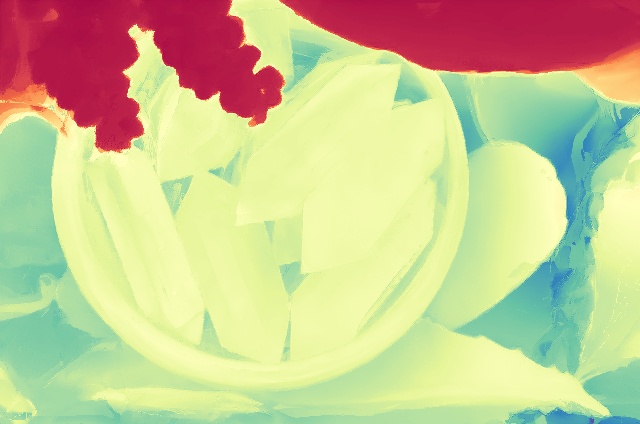}\\[0pt]
        \figline{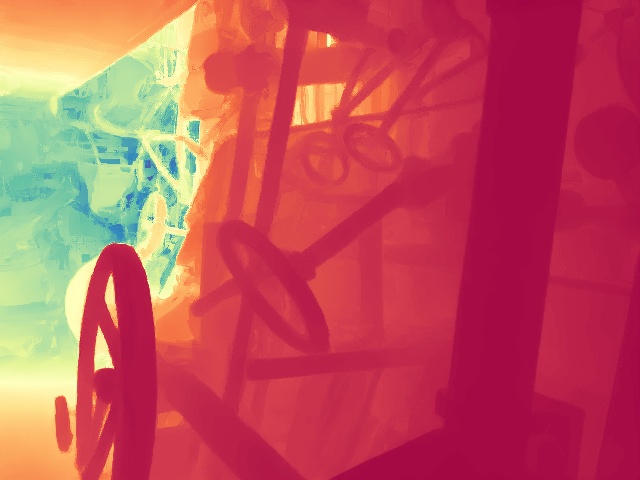}\\[0pt]
        \figline{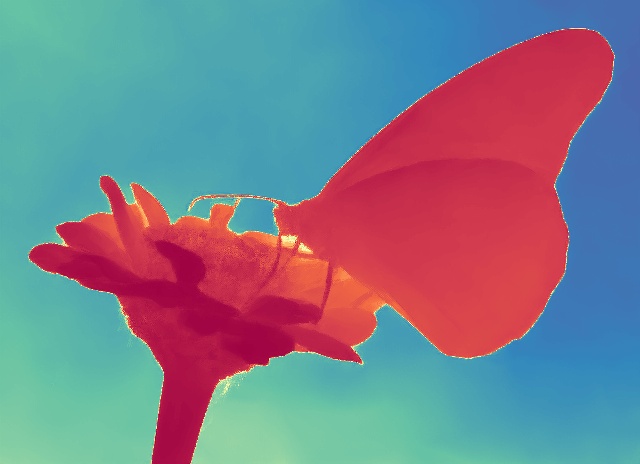}\\[0pt]
        \figline{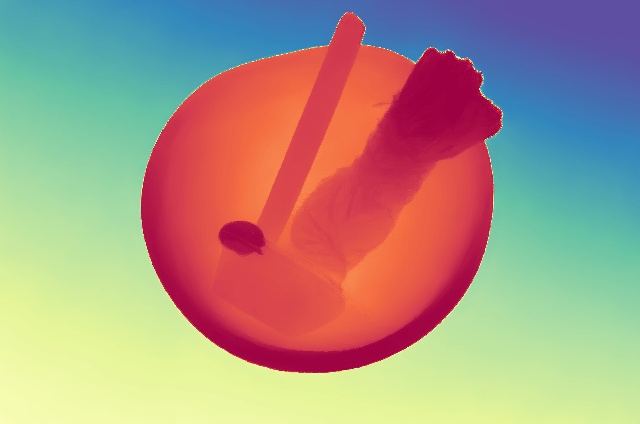}\\[0pt]
        \figline{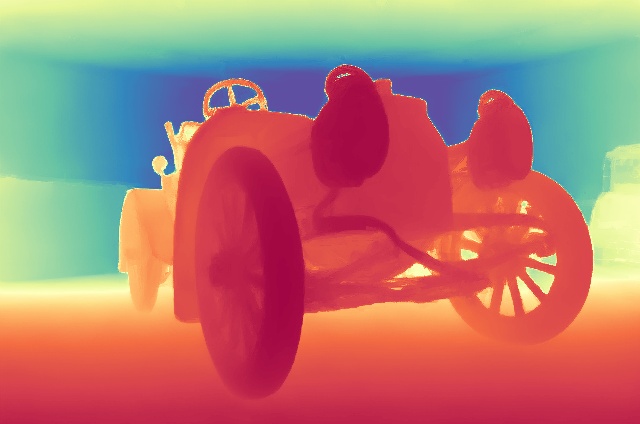}\\[0pt]
        \figline{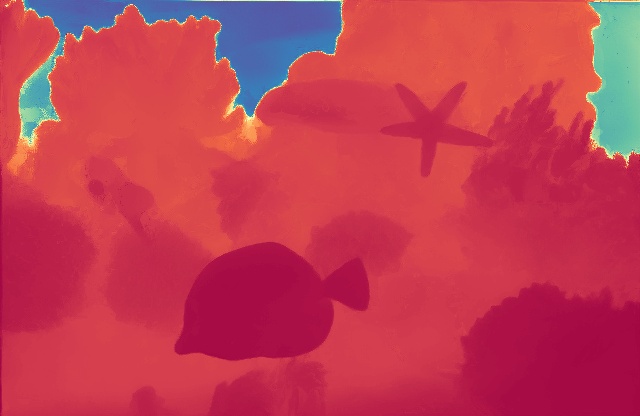}\\[0pt]
        \figline{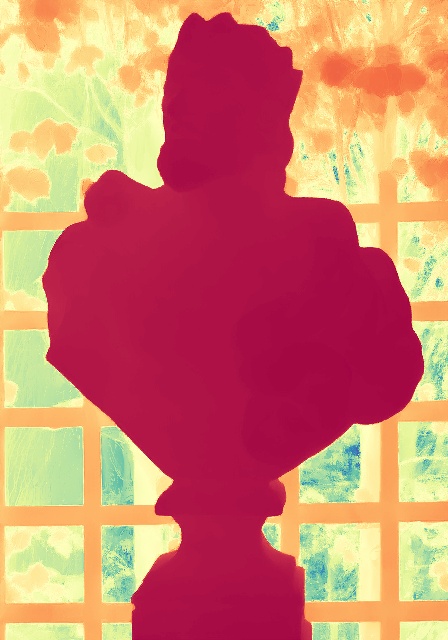}\\[0pt]
        RGB & Marigold (1, 1) & Marigold (50, 10) & Marigold + E2E FT & GeoWizard + E2E FT
    \end{tabularx}
    \caption{
        \textbf{Additional qualitative samples for depth estimation.}
        ``Marigold ($X$, $Y$)'' denotes Marigold using $X$ inference steps with an ensemble of size $Y$.
    }
    \label{fig:supp_qualitative_depth}
\end{figure*}

\begin{figure*}
    \centering
    \newcommand{\figblock}[2]{\begin{tikzpicture}\node[anchor=south west, inner sep=0] (img1) at (0,0) {\includegraphics[width=#1]{figures/raw/normal_640/#2.png}};\end{tikzpicture}}\newcommand{\figline}[1]{\includegraphics[width=0.19\textwidth]{figures/raw/rgb/#1_resized_640}&\figblock{0.19\textwidth}{#1_normalgold_1_1}&\figblock{0.19\textwidth}{#1_normalgold_50_10}&\figblock{0.19\textwidth}{#1_anggold}&\figblock{0.19\textwidth}{#1_ssiwiz}}
    \setlength{\tabcolsep}{0pt}
    \footnotesize
    \begin{tabularx}{\textwidth}{YYYYY}
        \figline{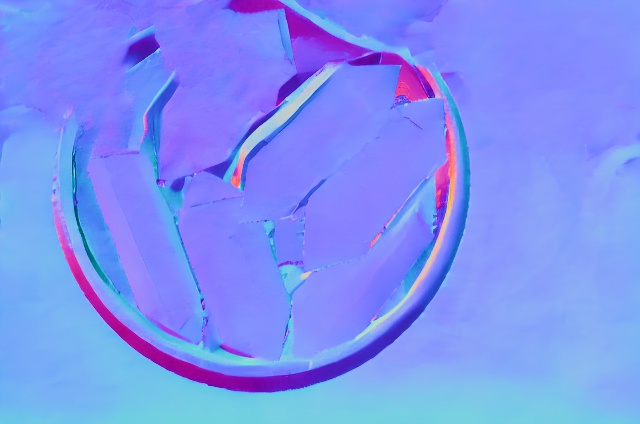}\\[0pt]
        \figline{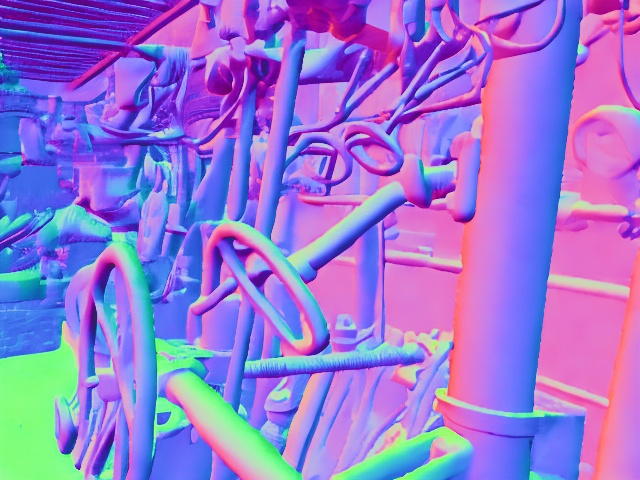}\\[0pt]
        \figline{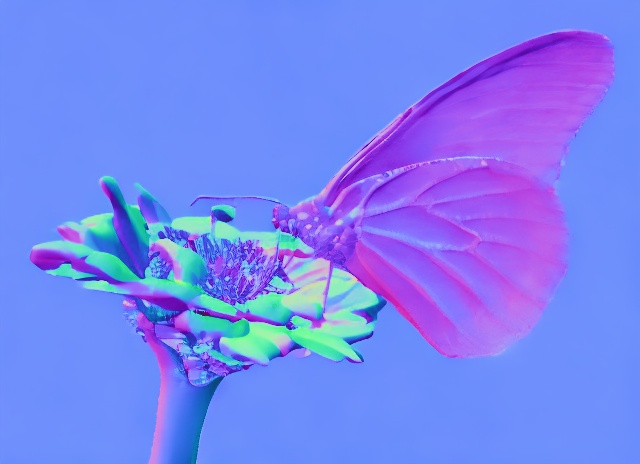}\\[0pt]
        \figline{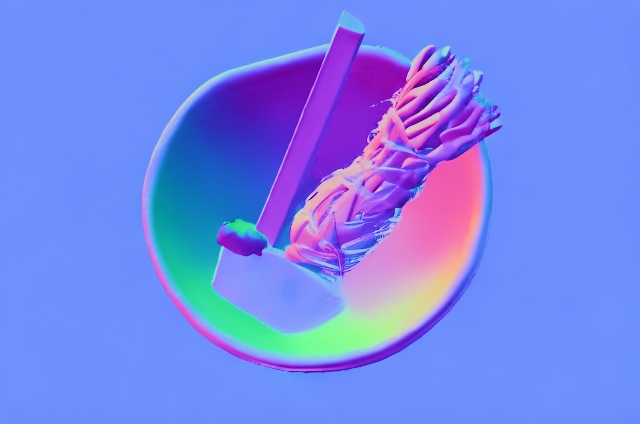}\\[0pt]
        \figline{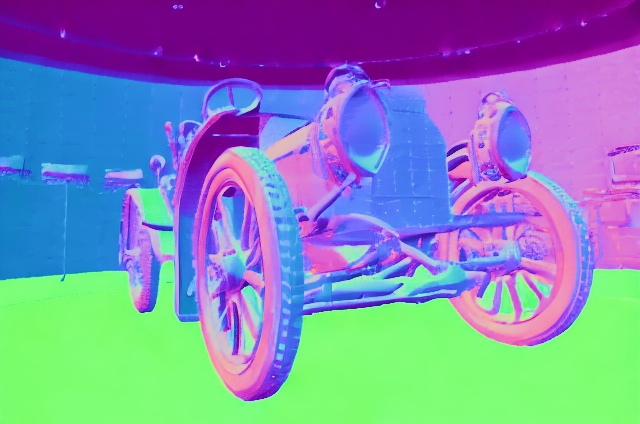}\\[0pt]
        \figline{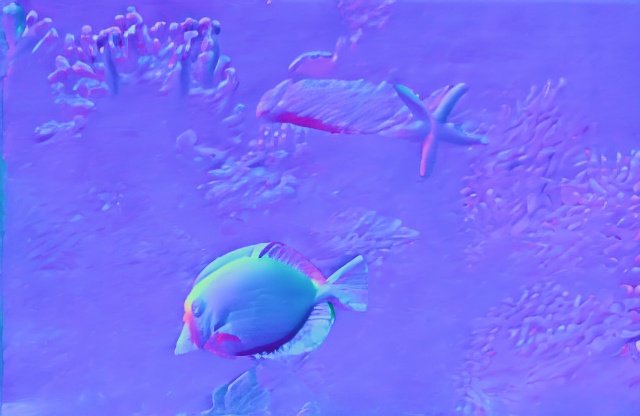}\\[0pt]
        \figline{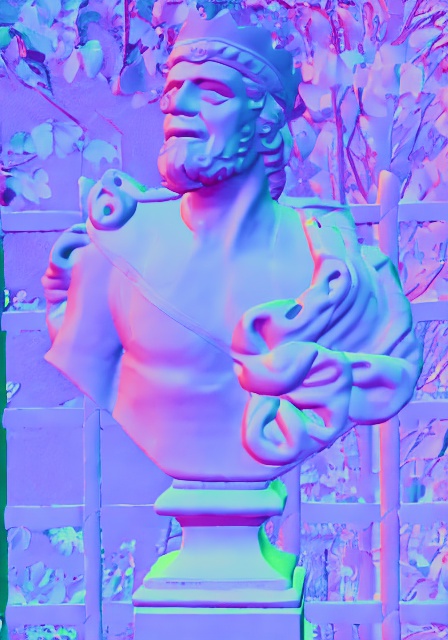}\\[0pt]
        RGB & Marigold (1, 1) & Marigold (50, 10) & Marigold + E2E FT & GeoWizard + E2E FT
    \end{tabularx}
    \caption{
        \textbf{Additional qualitative samples for normal estimation.}
        ``Marigold ($X$, $Y$)'' denotes Marigold using $X$ inference steps with an ensemble of size $Y$.
    }
    \label{fig:supp_qualitative_normals}
\end{figure*}

\paragraph{Further Qualitative Samples.}
\cref{fig:supp_qualitative_depth} and \cref{fig:supp_qualitative_normals} show qualitative results for depth and normals estimation, respectively, comparing Marigold~\cite{ke2023marigold} and the end-to-end fine-tuned models.
The fixed single-step model fails to produce sharp results, while the multi-step model exhibits noticeable over-sharpening and high-frequency noise artifacts (even after ensembling), particularly in the normals estimations.
In contrast, the end-to-end fine-tuned models do not exhibit these issues.

\section*{Addendum}

We were made aware of recent work by Xu~\etal~\cite{xu2024genpercept}.
Similar to us, they directly fine-tune Stable Diffusion in an end-to-end fashion, however, we arrive to this point in a very different way.
We initially discovered the issue with the DDIM scheduler, fixed this in Marigold, and in turn arrived to an end-to-end fine-tuning scheme that works for Marigold.
Surprisingly, our ablations showed that this also works well for direct fine-tuning of Stable Diffusion.
The main contribution of Xu~\etal is an approach to fine-tune Stable Diffusion (for a broader spectrum of tasks). However, even with additional modules on top, their method achieves lower scores than some of the baselines.
As such, these results might lead one to conclude that end-to-end fine-tuning is not a suitable alternative to multi-step, diffusion-based depth and normal estimation.
In contrast, our simple end-to-end fine-tuning setup \textit{does} outperform diffusion baselines, demonstrating that it is an effective and efficient alternative.

%% file: tbl/decoder_freezing.tex
\scriptsize
\setlength{\tabcolsep}{1.5pt}
\begin{tabularx}{\linewidth}{
l
YYc
YYc
YYc
YYc
YY
}

\toprule

\mr[l]{Decoder} &
\multicolumn{2}{c}{NYUv2 \cite{silberman2012nyuv2}} & &
\multicolumn{2}{c}{KITTI \cite{geiger2012kitti}} & &
\multicolumn{2}{c}{ETH3D \cite{schops2017eth3d}} & &
\multicolumn{2}{c}{ScanNet \cite{dai2017scannet}} & &
\multicolumn{2}{c}{DIODE \cite{vasiljevic2019diode}} \\

\cmidrule{2-3} 
\cmidrule{5-6} 
\cmidrule{8-9} 
\cmidrule{11-12} 
\cmidrule{14-15} 

& 
{\notsotiny AbsRel↓} & 
{\notsotiny $\delta$1↑} & &
{\notsotiny AbsRel↓} & 
{\notsotiny $\delta$1↑} & &
{\notsotiny AbsRel↓} & 
{\notsotiny $\delta$1↑} & &
{\notsotiny AbsRel↓} & 
{\notsotiny $\delta$1↑} & &
{\notsotiny AbsRel↓} & 
{\notsotiny $\delta$1↑} \\

\midrule

Frozen  & 
\textbf{5.2} & \textbf{96.6} & &
\textbf{9.6} & \textbf{91.9} & &
\textbf{6.2} &  95.9 & &
\textbf{5.8} & \textbf{96.2} & &
\textbf{30.2} &  \textbf{77.9} \\

Fine-tuned &
5.3 & 96.5 & &
\textbf{9.6} & \textbf{91.9} & &
\textbf{6.2} &  \textbf{96.0} & &
\textbf{5.8} & 96.1 & &
\textbf{30.2} &  77.7 \\

\bottomrule

\end{tabularx}

%% file: tbl/improve_geowizard_depth.tex
\footnotesize
\setlength{\tabcolsep}{2.5pt}
\begin{tabularx}{\textwidth}{
lccccc
YYc
YYc
YYc
YYc
YY
}

\toprule

\mr[l]{Method} &
\mr{Steps} &
\mr{Ensemble} &&
\mr{Inference \\time} &&
\multicolumn{2}{c}{NYUv2 \cite{silberman2012nyuv2}} & &
\multicolumn{2}{c}{KITTI \cite{geiger2012kitti}} & &
\multicolumn{2}{c}{ETH3D \cite{schops2017eth3d}} & &
\multicolumn{2}{c}{ScanNet \cite{dai2017scannet}} & &
\multicolumn{2}{c}{DIODE \cite{vasiljevic2019diode}} \\

\cmidrule{7-8}\cmidrule{10-11}\cmidrule{13-14}\cmidrule{16-17}\cmidrule{19-20}

& 
& 
& 
&
&
&
AbsRel↓ & 
$\delta$1↑ & &
AbsRel↓ & 
$\delta$1↑ & &
AbsRel↓ & 
$\delta$1↑ & &
AbsRel↓ & 
$\delta$1↑ & &
AbsRel↓ & 
$\delta$1↑ \\

\midrule

GeoWizard \cite{fu2024geowizard} &
50 &
10 &&
\SI{72}{\second}&&
\textcolor{noreproduce}{5.2} & 
\textcolor{noreproduce}{96.6} & &
\textcolor{noreproduce}{9.7} & 
\textcolor{noreproduce}{92.1} & &
\textcolor{noreproduce}{6.4} & 
\textcolor{noreproduce}{96.1} & &
\textcolor{noreproduce}{6.1} & 
\textcolor{noreproduce}{95.3} & &
\textcolor{noreproduce}{29.7} & 
\textcolor{noreproduce}{79.2} \\

\downrightarrow reproduced by us &
50 &
10 &&
\SI{72}{\second}&&
\underline{5.7} &
\textbf{96.2} & &
{14.4} &
{82.0} & &
\underline{7.5} & 
\underline{94.3} & &
\underline{6.1} & 
\underline{95.8} & &
\underline{31.4} & 
\underline{77.1} \\
\midrule

GeoWizard + DDIM fix &
1 &
1 &&
\textbf{\SI{254}{\milli\second}} &&
5.8 & 
\underline{96.1} & &
\underline{13.3} & 
\underline{84.7} & &
7.8 & 
\underline{94.3} & &
6.2 & 
95.7 & &
32.0 & 
76.0 \\

GeoWizard + E2E FT &
1 &
1 &&
\textbf{\SI{254}{\milli\second}} &&
\textbf{5.6} & \underline{96.1} & &
\textbf{9.8} & \textbf{91.4} & &
\textbf{6.3} &  \textbf{95.7} & &
\textbf{5.9} & \textbf{96.2} & &
\textbf{30.6} &  \textbf{77.9} \\

\bottomrule

\end{tabularx}

%% file: tbl/compare_depth_fm.tex
\footnotesize
\setlength{\tabcolsep}{2.5pt}
\begin{tabularx}{\textwidth}{
lccccc
YYc
YYc
YYc
YYc
YY
}

\toprule

\mr[l]{Method} &
\mr{Steps} &
\mr{Ensemble} &&
\mr{Inference \\time} &&
\multicolumn{2}{c}{NYUv2 \cite{silberman2012nyuv2}} & &
\multicolumn{2}{c}{KITTI \cite{geiger2012kitti}} & &
\multicolumn{2}{c}{ETH3D \cite{schops2017eth3d}} & &
\multicolumn{2}{c}{ScanNet \cite{dai2017scannet}} & &
\multicolumn{2}{c}{DIODE \cite{vasiljevic2019diode}} \\

\cmidrule{7-8}\cmidrule{10-11}\cmidrule{13-14}\cmidrule{16-17}\cmidrule{19-20}

& 
& 
& 
&
&
&
AbsRel↓ & 
$\delta$1↑ & &
AbsRel↓ & 
$\delta$1↑ & &
AbsRel↓ & 
$\delta$1↑ & &
AbsRel↓ & 
$\delta$1↑ & &
AbsRel↓ & 
$\delta$1↑ \\

\midrule

DepthFM \cite{gui2024depthfm} & 4 &  6 && \SI{1.67}{\second}&&
\textcolor{noreproduce}{ 6.5} & 
\textcolor{noreproduce}{ 95.6} & &
\textcolor{noreproduce}{ 8.3} & 
\textcolor{noreproduce}{ 93.4} & &
\textcolor{noreproduce}{ ---} & 
\textcolor{noreproduce}{ ---} & &
\textcolor{noreproduce}{ ---} & 
\textcolor{noreproduce}{ ---} & &
\textcolor{noreproduce}{ 22.5} & 
\textcolor{noreproduce}{ 80.0} \\

\downrightarrow reproduced by us & 4 &
6 &&
\SI{1.67}{\second}&&
6.9 &       
95.4 & &   
\underline{11.4} &     
88.1 & &   
6.5 & 
\textbf{96.2} & &
\underline{8.1} &    
92.5 & & 
\textbf{25.0} &   
78.3 \\   

DepthFM &
1 &
1 &&
\SI{132}{\milli\second}&&
7.5 &
95.0 & &
11.6 &
87.5 & &
6.7 &
\underline{96.0} & &
8.3 &
92.3 & &
\underline{25.3} &
\textbf{77.9} \\

\midrule

Marigold~\cite{ke2023marigold} + E2E FT 
&
1 &
1 &&
\textbf{\SI{121}{\milli\second}} &&
\textbf{5.2} & \textbf{96.6} & &
\textbf{9.6} & \underline{91.9} & &
\textbf{6.2} &  95.9 & &
\textbf{5.8} & \underline{96.2} & &
30.2 &  \textbf{77.9} \\

Stable Diffusion~\cite{rombach2021stablediffusion2} + E2E FT 
&
1 &
1 &&
\textbf{\SI{121}{\milli\second}} &&
\underline{5.4} & \underline{96.5} & &
\textbf{9.6} & \textbf{92.1} & &
\underline{6.4} &  95.9 & &
\textbf{5.8} & \textbf{96.5} & &
30.3 &  \underline{77.6} \\

\bottomrule

\end{tabularx}